\documentclass[10pt,twocolumn,numbers]{article}

\usepackage{graphicx}
\usepackage{amssymb}
\usepackage{amsthm,amsmath,dsfont,bbm}
\usepackage{mathtools}

\usepackage{lineno}

\usepackage[colorlinks=true, citecolor=blue]{hyperref}
\usepackage[linesnumbered,lined,ruled,commentsnumbered]{algorithm2e}
\usepackage{algcompatible}
\usepackage{wrapfig}
\usepackage[numbers,sort&compress]{natbib}
\usepackage{comment}
\usepackage{longtable}
\usepackage{multirow}
\usepackage{bbding}
  
\usepackage[bottom]{footmisc}
\usepackage{booktabs}
\usepackage{pifont}% http://ctan.org/pkg/pifont
\usepackage{float}
\usepackage{subfigure}
\usepackage{wrapfig}
\usepackage{framed,xcolor}
\usepackage{newfloat}
\usepackage[xcolor,framemethod=TikZ]{mdframed}
\usepackage{amsthm}
\usepackage{thmtools}
\usepackage{authblk}
\patchcmd{\thebibliography}{\section*{\refname}}{}{}{}
\usepackage{tikz}
\usetikzlibrary{arrows,shapes,chains,calc,patterns,decorations.pathreplacing}
\usepackage{stmaryrd}
\usepackage{fancyhdr}
\usepackage{subfigure}
\usepackage{comment}
\def\BibTeX{{\rm B\kern-.05em{\sc i\kern-.025em b}\kern-.08em
		T\kern-.1667em\lower.7ex\hbox{E}\kern-.125emX}}

\renewcommand{\headrulewidth}{2pt}

\newlength\FHoffset
\fancyheadoffset{\FHoffset}

\newlength\FHleft
\newlength\FHright

\newbox\FHline
\setbox\FHline=\hbox{\hsize=\paperwidth%
	\hspace*{\FHleft}%
	\rule{\dimexpr\headwidth-\FHleft-\FHright\relax}{\headrulewidth}\hspace*{\FHright}%
}

\newcommand{\RR}{\mathbb{R}}

\newcommand{\X}{\mathcal{X}}
\newcommand{\T}{\mathcal{T}}
\newcommand{\D}{\mathcal{D}}
\newcommand{\pp}{\mathrm{p}}
\newcommand{\ff}{\mathrm{f}}

\newtheoremstyle{theoremdd}% name of the style to be used
{\topsep}% measure of space to leave above the theorem. E.g.: 3pt
{\topsep}% measure of space to leave below the theorem. E.g.: 3pt
{\itshape}% name of font to use in the body of the theorem
{0pt}% measure of space to indent
{\fontfamily{cmss}\selectfont\bfseries}% name of head font
{.}% punctuation between head and body
{ }% space after theorem head; " " = normal interword space
{\thmname{#1}\thmnumber{ #2}\thmnote{ (#3)}}

\theoremstyle{theoremdd}

\mdfdefinestyle{myStyle}{roundcorner=5pt,backgroundcolor=brown!20,linecolor=red!40!black,linewidth=1.5pt}

\usepackage{titlesec}
\titleformat*{\section}{\fontfamily{cmss}\selectfont\large\bfseries\color{red!40!black}}
\titleformat*{\subsection}{\fontfamily{cmss}\selectfont\normalsize\bfseries\color{red!40!black}}
\titleformat*{\subsubsection}{\fontfamily{cmss}\selectfont\normalsize\color{red!40!black}}
\usepackage[left=0.69 in, right=0.69 in, top=1.1 in, bottom=1.3 in]{geometry}
\graphicspath{{./Figures/}}

\renewcommand\abstractname{\fontfamily{cmss}\selectfont\normalsize\bfseries\color{red!40!black}\textbf{Abstract}}

\renewenvironment{abstract}{%
	\centering\small
	\list{}{\leftmargin1.5cm \rightmargin\leftmargin}
	\item\relax
	
	\begin{mdframed}[]
		\item[\hskip\labelsep\scshape\abstractname.]%
	}{%
	\end{mdframed}
	\endlist \par\bigskip
}
\makeatletter
\patchcmd{\@maketitle}{\LARGE \@title}{\fontfamily{cmss}\selectfont\LARGE\color{red!40!black}\@title}{}{}
\makeatother

\graphicspath{{./Figures/}}

\usepackage{scalerel}
\usepackage{tikz}
\usetikzlibrary{svg.path}

\definecolor{orcidlogocol}{HTML}{A6CE39}
\tikzset{
	orcidlogo/.pic={
		\fill[orcidlogocol] svg{M256,128c0,70.7-57.3,128-128,128C57.3,256,0,198.7,0,128C0,57.3,57.3,0,128,0C198.7,0,256,57.3,256,128z};
		\fill[white] svg{M86.3,186.2H70.9V79.1h15.4v48.4V186.2z}
		svg{M108.9,79.1h41.6c39.6,0,57,28.3,57,53.6c0,27.5-21.5,53.6-56.8,53.6h-41.8V79.1z M124.3,172.4h24.5c34.9,0,42.9-26.5,42.9-39.7c0-21.5-13.7-39.7-43.7-39.7h-23.7V172.4z}
		svg{M88.7,56.8c0,5.5-4.5,10.1-10.1,10.1c-5.6,0-10.1-4.6-10.1-10.1c0-5.6,4.5-10.1,10.1-10.1C84.2,46.7,88.7,51.3,88.7,56.8z};
	}
}

\newcommand\orcidicon[1]{\href{https://orcid.org/#1}{\mbox{\scalerel*{
				\begin{tikzpicture}[yscale=-1,transform shape]
					\pic{orcidlogo};
				\end{tikzpicture}
			}{|}}}}

\begin{document}

%	\begin{frontmatter}
		
		
		\title{Incorporating Kinematic Wave Theory into a Deep Learning Method for High-Resolution Traffic Speed Estimation}
		
			

\author[1,2]{Bilal~Thonnam~Thodi$^{1,2}$, %\orcidicon{0000-0002-3927-4856}
	~Zaid~Saeed~Khan$^{\dagger1,2}$, %\orcidicon{0000-0003-2529-0000}
	~Saif~Eddin~Jabari$^{1,2}$, %\orcidicon{0000-0002-2314-5312}
	and~M\'onica~Men\'endez%$^{~1,2}$ %\orcidicon{0000-0001-5701-0523}
	}

\affil[1]{New York University Tandon School of Engineering, Brooklyn NY, U.S.A.}
\affil[2]{New York University Abu Dhabi, Saadiyat Island, P.O. Box 129188, Abu Dhabi, U.A.E.}
\affil[$^{\dagger}$]{Corresponding author. Email: \href{mailto:zaid.khan@nyu.edu}{zaid.khan@nyu.edu}}

\date{}

%\begin{mdframed}[style=myStyle]
%\end{mdframed}

\twocolumn[
\begin{@twocolumnfalse}
	
\maketitle	

\begin{abstract}
    We propose a kinematic wave-based Deep Convolutional Neural Network (Deep CNN) to estimate high-resolution traffic speed fields from sparse probe vehicle trajectories. We introduce two key approaches that allow us to incorporate kinematic wave theory principles to improve the robustness of existing learning-based estimation methods. First, we propose an anisotropic traffic kernel for the Deep CNN. The anisotropic kernel explicitly accounts for space-time correlations in macroscopic traffic and effectively reduces the number of trainable parameters in the Deep CNN model. Second, we propose to use simulated data for training the Deep CNN. Using a targeted simulated data for training provides an implicit way to impose desirable traffic physical features on  the learning model. In the experiments, we highlight the benefits of using anisotropic kernels and evaluate the transferability of the trained model to real-world traffic using the Next Generation Simulation (NGSIM) and the German Highway Drone (HighD) datasets. The results demonstrate that anisotropic kernels significantly reduce model complexity and model over-fitting, and improve the physical correctness of the estimated speed fields. We find that model complexity scales linearly with problem size for anisotropic kernels compared to quadratic scaling for isotropic kernels. Furthermore, evaluation on real-world datasets shows acceptable performance, which establishes that simulation-based training is a viable surrogate to learning from real-world data. Finally, a comparison with standard estimation techniques shows the superior estimation accuracy of the proposed method.
    
	\medskip
	
	\textbf{\fontfamily{cmss}\selectfont\color{red!40!black} Keywords}: 
	Traffic state estimation, traffic anisotropy, kinematic wave theory, convolutional neural networks, deep learning.
\end{abstract}
\bigskip
\end{@twocolumnfalse}
]

%\begin{keyword}
%	Automated vehicles \sep cellular Automata \sep conditional random fields \sep stochastic traffic modeling \sep traffic state estimation \sep trajectory reconstruction
%\end{keyword}
		
%\end{frontmatter}
	
	
%	\linenumbers
	

%%%%%%%%%%%%%%%%%%%%%%%%%%%%%%%%%%%%%%%%%%%%%%%%%%%%%%%%%%%%%%%%%%%%%%%%%%%%%%%%%%%
%%%%%%%%%%%%%%%%%%%%%%%%%%%%%%%%%%%%%%%%%%%%%%%%%%%%%%%%%%%%%%%%%%%%%%%%%%%%%%%%%%%
%%%%%%%%%%%%%%%%%%%%%%%%%%%%%%%%%%%%%%%%%%%%%%%%%%%%%%%%%%%%%%%%%%%%%%%%%%%%%%%%%%%

%%%%%%%%%%%%%%%%%%%%%%%%%%%%%%%%%%%%%%%%%%%%%%%%%%%%%%%%%%%%%%%%%%%%%%%%%%%%%%%%%%%
%%%%%%%%%%%%%%%%%%%%%%%%%%%%%%%%%%%%%%%%%%%%%%%%%%%%%%%%%%%%%%%%%%%%%%%%%%%%%%%%%%%
%%%%%%%%%%%%%%%%%%%%%%%%%%%%%%%%%%%%%%%%%%%%%%%%%%%%%%%%%%%%%%%%%%%%%%%%%%%%%%%%%%%

%%%%%%%%%%%%%%%%%%%%%%%%%%%%%%%%%%%%%%%%%%%%%%%%%%%%%%%%%%%%%%%%%%%%%%%%%%%%%%%%%%%
%%%%%%%%%%%%%%%%%%%%%%%%%%%%%%%%%%%%%%%%%%%%%%%%%%%%%%%%%%%%%%%%%%%%%%%%%%%%%%%%%%%
%%%%%%%%%%%%%%%%%%%%%%%%%%%%%%%%%%%%%%%%%%%%%%%%%%%%%%%%%%%%%%%%%%%%%%%%%%%%%%%%%%%

\section{Introduction}
\label{sec1}

Traffic management agencies use a variety of monitoring and control tools to ensure the safe and efficient operation of network road traffic. To meet their operational goals, agencies employ tools that identify disturbances and deploy effective control strategies in real time \cite{papageorgiou2004overview}. However, this requires accurate and timely knowledge of traffic conditions over the entire network, which is currently not possible given the limited sensory instrumentation in most (if not all) cities today. Fixed sensors are expensive and tend to be sparsely installed, offering limited spatial coverage. Data from mobile sensors are expected to become more widely available than data from point sensors, but remain extremely limited in practice; their sparsity is temporal \cite{ambhul2016mfd}. To address such data sparsity (spatially or temporally), we need appropriate mechanisms that fill the gaps in the traffic observations. These are known as \emph{traffic state estimation} (TSE) tools \cite{seo2017traffic}. TSE is a critical precursor to a number of real-time traffic control strategies with either conventional vehicles or a mix with connected and autonomous vehicles \cite{papageorgiou2004overview,li2021backpressure}. Such strategies include, but are not limited to, ramp metering, perimeter control, traffic signal control, and vehicle routing \cite{li2019position,yang2017perimeter,lin2021pay}. 
% Existing TSE methods are categorized and discussed below, along with their advantages and drawbacks.

Existing TSE approaches can be broadly divided into two categories: model-based and data-driven \cite{seo2017traffic}. The former approach adopts a mathematical model of traffic flow such as the first-order Lighthill-Whitham-Richards (LWR) model \cite{lighthill1955kinematic,richards1956shock} or one of its many higher-order extensions, like the Aw-Rascle-Zhang (ARZ) model \cite{aw2000resurrection,zhang2002non}. These methods assimilate flow model predictions with real-world observations using an exogenous filter (e.g., Ensemble Kalman filter) \cite{vanerp2020relflow,dakic2018smsestimate,fountoulakis2017highway,bekiaris2016highway,hoogen2012lagrang,nantes2016real,jabari2012stochastic,jabari2013gauss}. Traffic flow models ensure that estimates respect basic traffic principles. However, the models are based on simplifying assumptions of traffic physics that can lead to numerical bias when the assumptions are not met. Furthermore, approximation errors can arise from the data assimilation techniques used in TSE. For instance, it is common to linearize a non-linear flow model for the recursive estimation, and the approximations are poor around the capacity region \cite{seo2017traffic,hoogen2012lagrang}. Lastly, model-based methods require additional inputs (e.g., boundary conditions) which are difficult to obtain in real-time.

The other category of TSE approaches include data-driven/learning techniques, which build statistical/machine learning models from large volumes of (historical) traffic data. Some commonly employed tools include (predominantly) deep neural networks \cite{benkraouda2020traffic,yuhan2016dnn-speed}, support vector regression \cite{xiao2018speed}, principal component analysis \cite{li2013efficient}, and matrix factorization methods \cite{li2021nonlinearts}. The estimation results from data-driven methods are often reported to be more accurate than model-based approaches, but these methods also have shortcomings. Being purely data-driven, the models are agnostic to the physics of traffic flow and could lead to infeasible estimation results. These methods are also not often interpretable and lack robustness. More importantly, the generalizability of the models is often weak and depends on the training data distribution. 
% Perhaps most importantly, these methods lack interpretability; they are often referred to as ``black boxes'' and criticized by some traffic experts for the difficulty of obtaining insight from the building blocks of the models.

We aim to develop a methodology that incorporates the desirable features of both categories, namely the combination of domain knowledge with representational power. Such structured learning methods can ensure robust and interpretable estimation results, parsimonious model complexity, and reduced data requirements. Some recent works along these lines include \cite{yuan2021phygp,zhang2020hybrid,jabari2020sparse,jabari2019learning,kaidi2019queueest,jabari2018stochastic}. A common flavor in these approaches is to impose the physical constraints as cost function regularizers (i.e., as soft constraints) or derive the learning model architecture from physical principles. For instance, \cite{jabari2019learning,jabari2018stochastic} combine predictions of stochastic traffic flow model and limited probe vehicle data to infer vehicle trajectory distributions that are consistent with traffic physics. \cite{kaidi2019queueest} estimates queue lengths at signalized intersection as a solution to a convex optimization problem with queue propagation constraints guided by the kinematic wave theory of traffic flow. \cite{yuan2021phygp} uses the dynamical equations of macroscopic flow models to regularize a Gaussian Process regression model, which is efficient in handling sparse and noisy data.

In the context of deep learning, which is a more attractive choice for non-linear modeling, \cite{shi2021pnntse,pmlr2021barreau,aai2021shi,barreau2021physics,huang2020physics} approximate the solution of a macroscopic traffic flow model using deep neural networks and use the governing physical dynamical equations (in the form of PDE/ODE) as a regularizer in the cost function. They demonstrate that these physics informed regularizers reduce the space of feasible solutions and learn solutions that are consistent with the chosen traffic flow models under limited real-world data. However, this requires training deep neural networks for every instance of initial/boundary conditions, which is computationally expensive for real-time implementation.

% In the deep learning literature, there is a new paradigm called \textit{Physics-informed deep learning} \cite{raissi2019physics}, where the governing physical dynamical equations (in the form of PDEs or ODEs) are incorporated into the architecture or cost function of the DNN, acting as a regularizer. 
% This is already being explored in the context of TSE in . In these works, the authors demonstrate that these physics informed regularizers reduce the space of feasible solutions and learn solutions that are consistent with the chosen traffic flow models with limited data. %However, this is only possible with governing equations that are well-posed and models which are easy to integrate with neural networks. Microscopic traffic behaviors (which involve car-following or lane changing for example) are difficult to incorporate.

We propose an addition to this nascent literature on structured learning methods for TSE that incorporates traffic domain knowledge into learning models. Specifically, we propose a methodology to estimate high-resolution macroscopic traffic speed fields from limited probe vehicle measurements. We use a Deep Convolutional Neural Network (Deep CNN) as the learning model for estimation. The Deep CNN model takes as input sparse vehicle trajectory measurements and outputs a high-resolution speed field over a given space-time domain. The model is trained offline and can then be applied for real-time estimation. We incorporate traffic-specific features into the learning model in two ways, which are described below.

First, the na\"ive isotropic kernels in the Deep CNN model are modified to capture the wave propagation characteristics of free-flowing and congested traffic, in accordance with the kinematic wave theory (KWT) of traffic flow \cite{newell1993simplified}. We develop a Deep CNN with \emph{anisotropic kernels} designed to consider space-time inputs that are in the direction of feasible traffic waves, bounded by forward waves in free-flow and backward waves in congested traffic. As a result, we can significantly reduce the effective number of kernel parameters and hence the Deep CNN model complexity. Further, restricting the CNN to consider only the relevant spatio-temporal input points results in feasible and robust estimation of traffic shockwaves.

Second, we train our Deep CNN model using simulated traffic data. Apart from resolving the data availability issue, this approach allows us to take an empirical distribution of any desirable traffic flow model and use it to train the Deep CNN. The empirical distribution is a surrogate representation of the traffic physics underlying the simulation model. This is a broader approach to incorporate the governing physics as it is easier to generate data corresponding to complex traffic behaviors rather than integrating them into the model architecture as in existing physics-informed learning methods. We demonstrate this by training the Deep CNN model with data generated from a microscopic traffic simulator, which consists of behavioral car-following, lane-changing and gap-acceptance models, and then test it with real-world data having similar traffic characteristics. A natural trade-off of this approach is that the learning model does not capture the exact physical traffic dynamics, but can incorporate a wide range of complex traffic behaviors. Similar methods have been explored in the context of automated systems such as robotic controls and object detection, whereby researchers use high-fidelity simulators or synthetic data instead of real-world data to train deep neural network models \cite{abbeel2017domain,jonathan2018domainrand}.

To summarize, the contributions of this paper are:
\begin{enumerate}
	\item We develop an anisotropic kernel design for CNNs following the wave propagation characteristics of traffic flow. This could be applied to traffic state estimation, prediction, and data imputation. We also suggest an optimization procedure to learn the optimal weights for the anisotropic kernels.
	
	\item We propose to use simulated traffic data for fitting the anisotropic Deep CNN model and test its performance on real-world datasets.
	
	\item We demonstrate the use of the anisotropic Deep CNN model for speed field estimation at fine space-time resolutions ($10$ meters $\times$ $1$ second in our experiments) using limited input vehicle trajectories ($5\%$ probe vehicle penetration rates). We show sample estimations of real-world traffic data from multiple sources.
% 	We reconstruct high resolution ($10$ meters~$\times$~$1$ second) traffic speed fields from limited probe vehicle data (with $5\%$ penetration levels) using the anisotropic Deep CNN. We show sample estimations of real-world traffic data from multiple sources.
	
	\item We extend our estimation methodology to handle unknown probe vehicle penetration rates by introducing an ensemble version of our Deep CNN model.
\end{enumerate}

The rest of the paper is structured as follows. We present the estimation problem setting, the anisotropic kernel design, and the optimization procedure in Section \ref{sec2}. We then describe the training data generation and the training experiments in Section \ref{sec3}. In Section \ref{sec4}, we present estimation results, compare the anisotropic CNN with the na\"ive isotropic variant, discuss the transferability of the estimation model to real-world freeway traffic, and explore the sensitivity of the results to different probe vehicle penetration levels. Finally, we conclude the study in Section \ref{sec5}.

\section{Estimation Methodology}
\label{sec2}

\subsection{Traffic speed field estimation problem}
\label{sec2_1}

A space-time domain $\D = \X \times \T$ representing a given road section is discretized into homogeneous segments $x_i \subset \X$ and time intervals $t_i \subset \T$, such that $\cup_i x_i = \X$ and $\cup_i t_i = \T$. Let $V(x,t)$ denote the value of the macroscopic speed field in a cell $(x,t) \in \D$. We use the cell size $|x|$ closer to the length of a vehicle and $|t|$ in the order of seconds (smaller than what existing estimation methods use \cite{seo2017traffic}) to enable high-resolution speed field estimation. Probe vehicles (PVs) provide local speed measurements $\big\{V^{\pp}(x_i^\pp, t_i^\pp)\big\}$ for some cells in $\D$; we represent this partial information by the tensor $\mathbf{z}^{\pp}$. We assume sparse observation settings, where only a few cells (e.g., 5-10\%) in $\mathbf{z}^{\pp}$ have speed information. We denote by $\mathbf{z}^{\ff}$ a tensor of estimates of the complete speed field $V(x,t)$ for the entire space-time domain $\D$. The estimation problem can be formally stated as learning a mapping function $g: \mathbf{z}^{\pp} \mapsto \mathbf{z}^{\ff}$.

The speed field $V(x,t)$ in each cell of the input tensor $\mathbf{z}^\pp$ is encoded using a three-dimensional RGB array (with domain $\{1,\hdots,255\}$) instead of a one-dimensional speed value. This is to differentiate cells occupied with a stopped vehicle (i.e., with $V(x,t)=0$) from empty cells. The output tensor $\mathbf{z}^{\ff}$ represents the complete macroscopic speed field over the domain $\D$ and can be encoded using the one-dimensional speed values. Thus, we have, $\mathbf{z}^\pp \in \{1,\hdots,255\}^{|\X| \times |\T| \times 3}$ and $\mathbf{z}^\ff \in \RR_{\ge0}^{|\X| \times |\T|}$.

\subsection{Deep Convolutional Neural Network (Deep CNN) model for estimation}
\label{sec2_2}

\begin{figure}[!thb]
	\centering
	\includegraphics[width=0.5\textwidth]{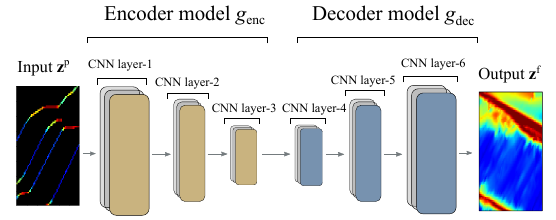}
	\caption{The architecture of the Deep CNN (speed reconstruction model).}
	\label{fig1:cnn}
\end{figure}

We use a Deep CNN model similar to the one in \cite{benkraouda2020traffic} to represent the mapping function $g$. The model architecture is shown in Fig. \ref{fig1:cnn}. It comprises an encoder $g_\mathrm{enc}$ and a decoder $g_\mathrm{dec}$, each consisting of three CNN layers. Each CNN layer is composed of a 2D convolution operation, a non-linear activation operation called ReLU (Rectified Linear Unit), and a down-sampling operation called max-pooling (up-sampling operation called nearest neighbor in case of $g_\mathrm{dec}$). As shown in Fig.~\ref{fig1:cnn}, the successive CNN layers of $g_\mathrm{enc}$ have reduced spatio-temporal widths and the successive CNN layers of $g_\mathrm{dec}$ have increased spatio-temporal widths. The Deep CNN model takes the input $\mathbf{z}^\pp$, passes it through the hierarchical convolution layers, and outputs the estimated speed field $\mathbf{z}^\ff$.

Unlike other neural network architectures, CNNs have proven to be effective in learning spatial data (e.g., images, video, etc.), which is useful for our application since the space-time diagram reflects spatial data. The CNN model has two properties favorable for learning macroscopic traffic features: local connectivity and parameter sharing. The former assumes the traffic speed fields are locally correlated, and the latter implies local traffic features can occur anywhere in the space-time plane, i.e., they are space-time invariant. Furthermore, the specific encoder-decoder structure (bottleneck formation) of the model shown in Fig.~\ref{fig1:cnn} can efficiently handle the sparse nature of the model input \cite{benkraouda2020traffic}.

The discrete convolution using local kernels in the CNN forms the basis of traffic speed field estimation. In a given CNN layer $l$, a convolution operation calculates the activation in a cell $(x, t)$ as a weighted sum of cell activations observed in the previous layer $(l-1)$:
\begin{multline}
	\mathbf{z}^{(l)} (x, t, \chi) = \mathbf{z}^{(l-1)}(\cdot,\cdot,\chi) \ast \Theta^{(l)}(\cdot,\cdot) \\ = \sum_{(x_j, ~t_j) \in ~I_\mathrm{iso}} \mathbf{z}^{(l-1)}\left( x_j, t_j, \chi \right) \Theta^{(l)} \left( x_j, t_j \right),
	\label{eqn1:conv}
\end{multline}
where $\mathbf{z}^{(l)} (x, t, \chi)$ is the feature map value in layer $l$ associated with cell $(x,t)$ and color channel $\chi\in\{1,2,3\}$, $\Theta^{(l)}(\cdot,\cdot) \in \RR^{|\X| \times |\T|}$ is the kernel (matrix), which is identical for all cells. $\Theta^{(l)} \left( x_j, t_j \right)$ on the right-hand side, an element of the kernel matrix, determines the extent to which neighboring cell $(x_j,t_j) \in I_\mathrm{iso}$ is correlated with the subject cell $(x,t)$. Hereafter, we simply write $\Theta$ to represent the entire kernel, and drop the `$(\cdot,\cdot)$'.

The feature map value in cell $(x,t)$ can be considered as equivalent to (or some function of) the speed field $V(x,t)$ in that cell. Then, operation \eqref{eqn1:conv} simply says: the speed in cell $(x,t)$ is a weighted interpolation of speeds observed in its immediate surrounding cells. The extent of local cell influence $I_\mathrm{iso}$ is depicted visually on the space-time plane in Fig. \ref{fig2:conv}(a). Each kernel in a layer $l$ represents a different weighting function; together, the kernels learn to identify different traffic features.

\subsection{Anisotropic kernel design for Deep CNN}
\begin{figure}[!tbh]
	\centering
	\resizebox{0.48\textwidth}{!}{%
	   \includegraphics[width=0.50\textwidth]{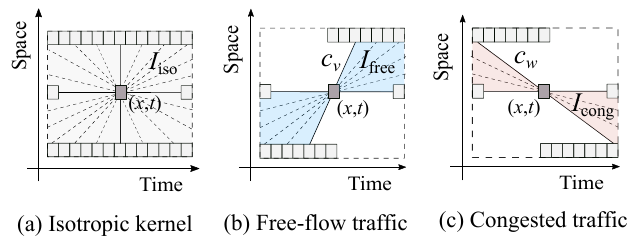}}
	\caption{Space-time correlations modeled by the isotropic kernel of the convolution operation, and that in the real traffic (free-flow and congested).}
	\label{fig2:conv}
\end{figure}

The isotropic kernel shown in Fig. \ref{fig2:conv}(a) says that the speed in cell $(x,t)$ is correlated with the speeds observed \textit{anywhere} in the shaded rectangular region $I_{\mathrm{iso}}$. This assumes that a speed variation (such as that caused by slowdowns or speed-ups) at $(x,t)$ can propagate at unbounded velocities in the space-time plane. However, this is not true in real traffic. In real traffic, (i) the speed/density variations propagate at finite velocities that are less than or equal to the free-flowing vehicle speed, and (ii) vehicles respond (predominantly) to frontal stimuli with a delay (approximately equal to the reaction time of driver). The former condition is called hyperbolicity and the latter is called anisotropy. Hyperbolicity is a necessary but not sufficient condition for anisotropy in traffic flow models \cite{hoogendorn2013ani,daganzo1995req,treiber2013traffic}. The use of local kernels (i.e., kernel dimensions $\ll$ the dimension of space-time plane) captures the hyperbolicity property, whereas anisotropy can be captured by modifying the kernel shape as discussed below.

The actual propagation velocity of speed variations depends on the traffic state (i.e., speed or density). We assume that traffic at any point in the space-time plane is either in free-flow or is congested in relation to the Fundamental Diagram. There are different traffic conditions associated with free-flow and congestion, respectively. Then, a speed variation in cell $(x,t)$ propagates downstream (i.e., in the direction of traffic) in free-flowing traffic and upstream (i.e., in the opposite direction of traffic) in congested traffic. This is an empirically and theoretically established feature of traffic \cite{newell1993simplified,treiber2002filter,trieber2011filter,daganzo2005variational,daganzo2005var_bott}. Thus, the extent of the space-time plane correlated with cell $(x,t)$ depends on whether the traffic state is free-flow or congested. The respective correlated regions are shown in Fig. \ref{fig2:conv}(b) and Fig. \ref{fig2:conv}(c) as shaded areas $I_{\mathrm{free}}$ and $I_{\mathrm{cong}}$. 

The regions $I_{\mathrm{free}}$ and $I_{\mathrm{cong}}$ are bounded by the free flow traffic speed $c_v$ and the backward shockwave speed $c_w$ \cite{hoogendorn2013ani,treiber2013traffic}, respectively. The speed in cell $(x,t)$ influences the region $I_\mathrm{free}$ downstream, and the region $I_\mathrm{cong}$ upstream. Likewise, the regions $I_\mathrm{free}$ upstream and $I_\mathrm{cong}$ downstream influence the speed in cell $(x,t)$. In summary, the speed predicted in cell $(x,t)$ is correlated with the speeds observed anywhere in $I_\mathrm{free} \cup I_\mathrm{cong}$. We use this knowledge of space-time correlations in designing an alternate and causally \textit{correct} kernel (in the traffic sense) for the Deep CNN model in Fig. \ref{fig1:cnn}. We refer to this as the \textit{anisotropic kernel}, and represent it by the tensor $\Theta_{\mathrm{ani}} = [\Theta_{\mathrm{ani}}^{(l)}]_l$. The corresponding convolution operation is slightly modified from \eqref{eqn1:conv} as,
\begin{multline}
	\mathbf{z}^{(l)} (x, t, \chi) = \mathbf{z}^{(l-1)}(\cdot,\cdot,\chi) \ast \Theta_{\mathrm{ani}}^{(l)} \\ = \sum_{(x_j, ~t_j) \in ~I_\mathrm{ani}} \mathbf{z}^{(l-1)}\left( x_j, t_j, \chi \right) \Theta_{\mathrm{ani}}^{(l)} \left( x_j, t_j \right),
	\label{eqn2:conv2}
\end{multline}
where the effective influence region is defined as $I_\mathrm{ani} := I_\mathrm{free} \cup I_\mathrm{cong}$. This way, we direct the convolution operator to consider only that portion of the space-time plane which is relevant for the speed interpolation according to traffic physics.

\begin{figure}[!tbh]
	\centering
	\includegraphics[width=0.5\textwidth]{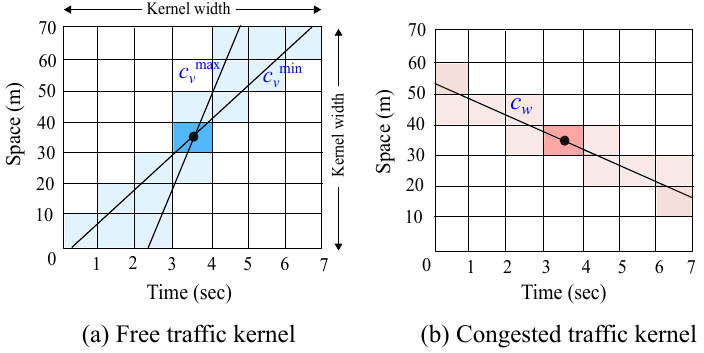}
	
	\includegraphics[width=0.5\textwidth]{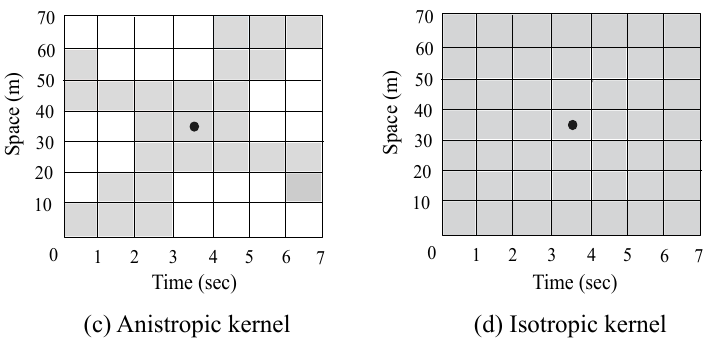}
	
	\caption{The anisotropic kernel design for a $7 \times 7$ CNN kernel (width = 7). The na\"ive isotropic kernel is also shown here for comparison. Parameters used: $c_v^{\mathrm{max}}=100$ kmph, $c_v^{\mathrm{min}}=60$ kmph and $c_w=18$ kmph.}
	\label{fig3:ker}
\end{figure}

In this paper, we propose a specific anisotropic kernel design, whose influence region is further restricted, motivated by empirical observations: (i) congested traffic has a very narrow range of wave propagation velocities (such that they can be regarded as almost constant), and (ii) free-flow traffic wave propagation velocities are limited within the maximum and minimum desired vehicle speeds \cite{newell1993simplified,treiber2002filter,trieber2011filter,treiber2013traffic,hoogendorn2013ani}. 
% \textcolor{blue}{The former observation is especially true for freeway traffic. For urban traffic with queuing at traffic signals, a wide range of backward wave speeds would be more appropriate.} 
The anisotropic kernel design to replace the isotropic kernel (from Fig. \ref{fig1:cnn}) is illustrated in Fig. \ref{fig3:ker}. We create two kernels, one each for free-flowing and congested traffic. The influence region $I_\mathrm{free}$ contains all the cells passing and bounded between the  maximum ($c_v^{\mathrm{max}}$) and minimum ($c_v^{\mathrm{min}}$) desired vehicle speeds. This is relevant for heterogeneous traffic where the desired speed distribution has a wide range. The free-flow traffic kernel is shown in Fig. \ref{fig3:ker}(a). The influence region for congested traffic, $I_\mathrm{cong}$, contains only those cells passing through the backward propagating shockwave speed $c_w$; see Fig. \ref{fig3:ker}(b). The proposed anisotropic kernel is a superposition of the free-flow and congested kernel. This is shown in Fig. \ref{fig3:ker}(c). The corresponding isotropic kernel is shown in Fig. \ref{fig3:ker}(d) for comparison. One can see that the anisotropic design requires $50\%$ fewer parameters than its isotropic variant for a $7 \times 7$ kernel. 

In summary, our proposed anisotropic kernel design takes three input parameters $\{c_v^{\mathrm{max}}, c_v^{\mathrm{min}}, c_w\}$, whose values depend on the traffic characteristics of the road section. The proposed design aims to learn a broad range of forward propagation speeds and a narrow range of backward propagation speeds. Using a wide distribution for propagation speeds can simultaneously handle different road classes, e.g., highways with different speed limits and arterials. The variability in the free-flow speeds, in addition to capturing differences in speed limits, allows our kernels to capture a variety of kinematic wave speeds as combinations of free-flow waves and backward waves.

\begin{figure}[!tbh]
	\centering
	\includegraphics[width=0.30\textwidth]{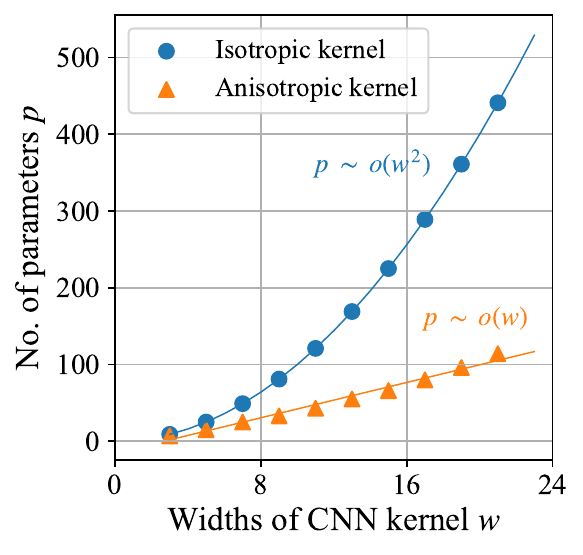}
	\caption{Number of model parameters for different widths of isotropic and anisotropic kernel.}
	\label{fig:kersize}
\end{figure}

A practical benefit of the proposed anisotropic kernel design is the significant reduction in the model complexity of the Deep CNN model. Model complexity here refers to the total number of model parameters, and depends on the widths and depths of CNN kernels. We quantify the parameter requirements for isotropic and anisotropic kernels as a function of kernel widths in Fig. \ref{fig:kersize}. The number of parameters scales \emph{linearly} for anisotropic kernels as opposed to \emph{quadratically} for isotropic kernels. This implies that anisotropic kernels result in a simpler, lower complexity CNN model which is easier to compute and optimize as compared to its isotropic counterpart. This scaling advantage is realized for higher kernel widths which naturally occur for larger problem sizes. We show these benefits experimentally in Section \ref{sec4} as we compare the model complexity requirements for different road network sizes.

We finally note that the proposed anisotropic kernel design is similar to Treiber and Helbing's adaptive smoothing method for speed interpolation \cite{treiber2002filter} except that: (a) we consider a range of wave propagation speeds in free-flowing traffic instead of a constant value, (b) the weights in the kernel are not set apriori as in \cite{treiber2002filter} but learned from data, and (c) the actual speed predicted is a combination of several anisotropic kernels as opposed to a single anisotropic kernel.

\subsection{Learning anisotropic kernels}

We use anisotropic kernels in all layers of the Deep CNN model in Fig. \ref{fig1:cnn}. The optimal weights $\Theta_{\mathrm{ani}}^*$ for the anisotropic kernel are obtained from the following constrained optimization problem:
\begin{equation}
	\Theta_{\mathrm{ani}}^* := \underset{\Theta \in \RR^{|\X| \times |\T|\times L}}{\arg \min} \big\{ \mathcal{L} \left( \mathbf{z}^\ff, g\left( \mathbf{z}^\pp, \Theta \right) \right): ~ \mathbbm{1}_\mathrm{ani} \odot \Theta   = \mathbf{0}  \big\},
	\label{eqn3:keropt}
\end{equation}
where $g(\mathbf{z}^\pp, \Theta): \{0,\hdots,255\}^{|\X| \times |\T| \times 3} \rightarrow \RR_{\ge 0}^{|\X| \times |\T|}$ is the mapping function (i.e., the Deep CNN) with the kernel parameterization $\Theta$ made explicit (i.e., $g$ performs the mapping $\mathbf{z}^\pp \mapsto \mathbf{z}^\ff$), $\mathbbm{1}_\mathrm{ani}$ is a binary tensor of the same dimension as $\Theta$ with values of 0 for cells corresponding to the anisotropic influence cell region $I_\mathrm{ani}$, e.g., the shaded cells in Fig.~\ref{fig3:ker}(c), and 1 elsewhere ($\odot$ is the Hadamard product). The loss function $\mathcal{L}$ captures any discrepancies between the estimated and true speed fields, e.g., the squared $\ell_2$ distance (the squared error): 
\begin{equation}
	\mathcal{L} \left( \mathbf{z}^\ff, g\left( \mathbf{z}^\pp, \Theta \right) \right) = \left\| \mathbf{z}^\ff - g\left( \mathbf{z}^\pp, \Theta \right) \right\|_2^2. \label{eqn:loss}
\end{equation} 

The constrained optimization problem \eqref{eqn3:keropt} can be solved using iterative schemes which can handle feasibility constraints, such as the projected gradient descent. In each iteration $i$, the updates are calculated as follows:
\begin{equation}
	\Theta_{\mathrm{ani}}^{i+1} := \mathsf{P}_{I_{\mathsf{ani}}}\big( \Theta_{\mathrm{ani}}^{i} - \gamma^{i} G(\Theta_{\mathrm{ani}}^{i}) \big),
\end{equation}
where $\gamma_{i} > 0$ is the step size (or learning rate) in iteration $i$ and $G(\Theta_{\mathrm{ani}}^{i})$ is a gradient tensor (descent direction) at $\Theta_{\mathrm{ani}}^{i}$. The operator $\mathsf{P}_{I_{\mathrm{ani}}}$ assigns zeros to elements of $\Theta_{\mathrm{ani}}^{i} - \gamma^{i} G(\Theta_{\mathrm{ani}}^{i})$ corresponding to cells that lie outside of $I_{\mathrm{ani}}$, thereby ensuring feasibility of the solutions.

\section{Data and Training}
\label{sec3}

As mentioned earlier, we use simulated traffic data consisting of different traffic conditions for training the anisotropic Deep CNN model. In the following, we describe the data used for training and evaluating the model.

\subsection{Training data generation}

To generate data for training the CNN model, we simulate a freeway segment using the Vissim microscopic traffic simulator. The simulated segment corresponds to the \textit{E-22 Abu Dhabi-Al Ain road, UAE} (2 miles in length and 3 lanes wide), and includes an entry and exit ramp to a nearby suburban region. The simulation model is calibrated with general traffic behavior, for instance, prioritizing through movements, appropriate yielding gaps for on-ramp vehicles, and minimum gap for lateral movements. A wide distribution of desired vehicle speeds (ranging from $60-100$ kmph) is used to produce different free-flow wave propagation speeds as is the case for heterogeneous traffic.

\begin{figure}[!hbt]
	\begin{center}
		\includegraphics[width=0.40\textwidth]{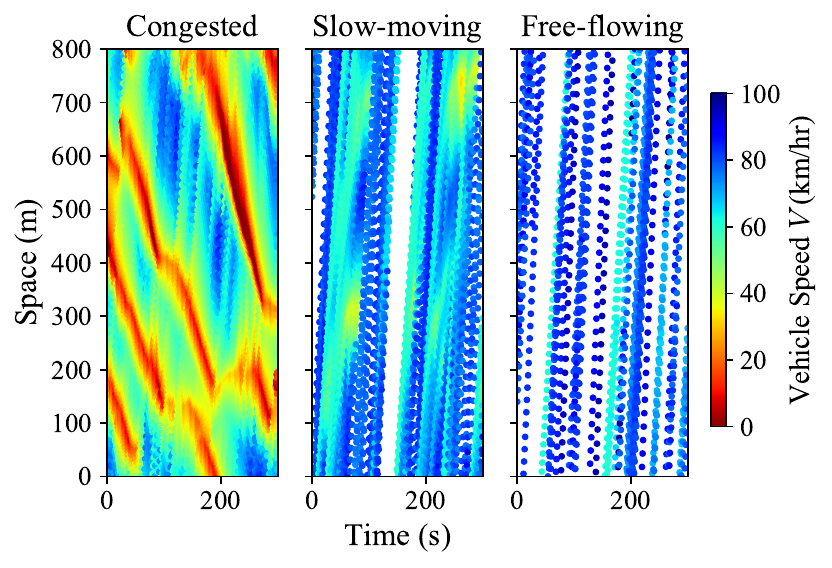} \newline
		{\small (a) Space-time-speed contour plot.}
		\vspace{0.1in}
		
		\includegraphics[width=0.35\textwidth]{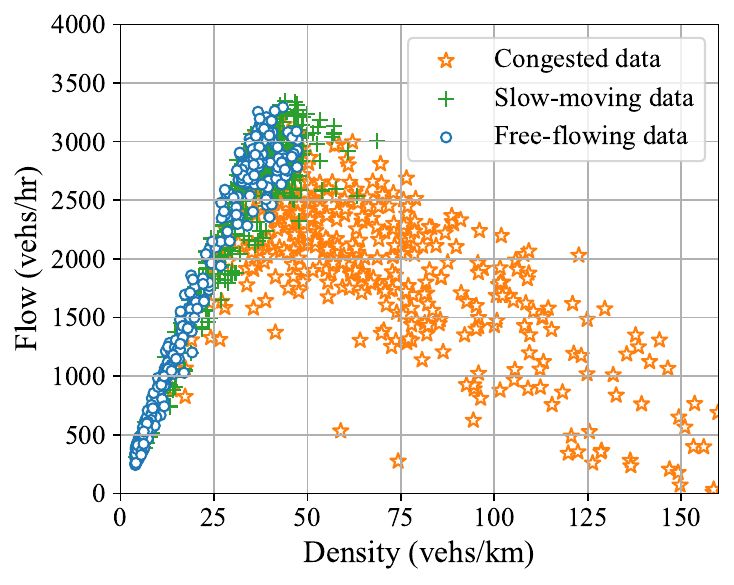} \newline
		{\small (b) Flow-density scatter plot.}
	\end{center}
	\caption{Visualization of the richness or the traffic features contained in the simulated training dataset (300-second snapshot).}
	\label{fig:sim_data}
\end{figure}

We simulate three traffic scenarios with different input vehicle demand profiles on the freeway segment: 800-1200 vehs/hr, 2400-3000 vehs/hr, and 4200-5400 vehs/hr. We used these demand profiles to replicate distinct traffic conditions on the simulated freeway, namely free-flowing, slow-moving (moderately congested), and (heavily) congested traffic. We used on-ramp inflows that constitute 15-20\% of the total freeway flows. Each traffic scenario is simulated for 2 hrs and the vehicle trajectory data for an 800 m homogeneous section on the freeway is recorded. The trajectory data corresponding to three traffic scenarios and their traffic dynamics are summarized in Fig.~\ref{fig:sim_data}.

Fig.~\ref{fig:sim_data}(a) shows a $300$ second snapshot of vehicle trajectories for the three simulated traffic scenarios. One can note the backward and forward propagating waves due to the stop-and-go, slow-moving, and heterogeneous free-flowing traffic (respectively) in Fig.~\ref{fig:sim_data}(a). The anisotropic kernel is designed based on the range of wave propagation speeds seen in these plots. Fig.~\ref{fig:sim_data}(b) is a flow-density scatter-plot of the three scenarios. Together, these figures show the richness of traffic states contained in the training data. 
% Later, we will use the flow-density scatter plot as a way to compare the traffic dynamics observed in the simulated and real-world traffic data.

\subsection{Definition of macroscopic speed field}

An important auxiliary task is to define the ``true'' speed field which the Deep CNN model uses as the ``ground truth'' for evaluating the quality of the estimation. This is achieved by \emph{translating}
the set of \emph{all} vehicle trajectories (not just PVs) into a speed field $V(x,t)$. The commonly used generalized definition of macroscopic speeds \cite{eddie1963traffic} results in $V(x,t) = 0$ for some cells due to the fine mesh size we use. Therefore, we propose a simple interpolation method for this purpose instead. Our method interpolates the speeds over the road cells at a fixed time according to:
\begin{equation}
	V(x,t) = 
	\begin{cases}
		V_\mathrm{up} \left( \frac{d_\mathrm{dn}}{d_\mathrm{up}+d_\mathrm{dn}} \right) + V_\mathrm{dn} \left( \frac{d_\mathrm{up}}{d_\mathrm{up}+d_\mathrm{dn}} \right), & \text{if } d_\mathrm{up} < l_\mathrm{up} \\ & \mkern-18mu \text{and } d_\mathrm{dn} < l_\mathrm{dn}  \\
		
		V_\mathrm{up} \left( 1- \frac{d_\mathrm{up}}{l_\mathrm{up}} \right) + V_\mathrm{max} \left( \frac{d_\mathrm{up}}{l_\mathrm{up}} \right), & \text{if } d_\mathrm{up} < l_\mathrm{up} \\ & \mkern-18mu \text{and } d_\mathrm{dn} > l_\mathrm{dn}  \\
		
		V_\mathrm{dn} \left( 1- \frac{d_\mathrm{dn}}{l_\mathrm{dn}} \right) + V_\mathrm{max} \left( \frac{d_\mathrm{dn}}{l_\mathrm{dn}} \right), & \text{if } d_\mathrm{up} > l_\mathrm{up} \\ & \mkern-18mu \text{and } d_\mathrm{dn} < l_\mathrm{dn}  \\
		
		V_{\mathrm{max}}, & \text{otherwise}, \\

% 		V_\mathrm{up} w + V_\mathrm{dn} (1-w), & \text{if } \left( x-x_\mathrm{up} \right) < l_\mathrm{up} \\ & \text{and} \left( x_\mathrm{dn} - x \right) < l_\mathrm{dn}  \\
% 		V_\mathrm{up} w + V_\mathrm{max} (1-w), & \text{if } \left( x-x_\mathrm{up} \right) < l_\mathrm{up} \\ & \text{and} \left( x_\mathrm{dn} - x \right) > l_\mathrm{dn} \\
% 		V_\mathrm{dn} w  + V_\mathrm{max} (1-w), & \text{if } \left( x-x_\mathrm{up} \right) > l_\mathrm{up} \\ & \text{and} \left( x_\mathrm{dn} - x \right) < l_\mathrm{dn} \\        
% 		V_{\mathrm{max}}, & \text{otherwise}, \\
	\end{cases}
	\label{eqn4:speed}
\end{equation}
where $V_\mathrm{max}$ is the highest free-flow speed (or speed limit of the highway section), $V_\mathrm{dn}$ (resp. $V_\mathrm{up}$) is the speed of the downstream (resp. upstream) vehicle, $d_\mathrm{dn}$ % = x_\mathrm{dn} - x$ 
(resp. $d_\mathrm{up}$) % = x - x_\mathrm{up}$
is the distance between the cell $(x,t)$ and the cell containing the downstream (resp. upstream) vehicle, and $l_\mathrm{dn}$ (resp. $l_\mathrm{up}$) is the length of spatial interaction downstream (resp. upstream) of $(x,t)$.

Equation \eqref{eqn4:speed} can be understood as follows: the speed field $V(x,t)$ in cell $(x,t)$ is a weighted combination of the speeds upstream and downstream of the cell. The speed $V_\mathrm{dn}$ of the vehicle downstream of $(x,t)$ has an effect only if it is within the downstream interaction range $l_\mathrm{dn}$ from $(x,t)$; otherwise, its value is replaced by the maximum highway speed $V_\mathrm{max}$ (and analogously for the upstream vehicle). The weights of the upstream and downstream components are proportional to the proximity of the respective interactions. The spatial interaction lengths are chosen to satisfy $l_\mathrm{up} < l_\mathrm{dn}$, to reflect the asymmetrically greater influence of frontal interaction.

\subsection{Training procedure}

%We use the following space-time discretization parameters for the training samples: $X = 800$ m, $T = 60$ s, $x = 10$ m, and $t = 1$ s. The input-output samples required for training are generated as described below. 
The simulation output for each scenario is 7200 seconds of trajectory data for each of the three lanes. We first map the trajectories from a single lane onto a space-time plane to form an input and output frame of dimension $80 \times 7200$ (i.e., the mesh size is $10$ m $\times$ $1$ s). The PV trajectories for the input frame are selected at random using a $5\%$ sampling rate. The output frame that forms the ground truth speed field is generated using the interpolation procedure described in eq.~\eqref{eqn4:speed}. We then extract samples of the input ($\mathbf{z}^\pp$) and output tensors ($\mathbf{z}^\ff$) from the input and output frames respectively, using a $80 \times 60$ sliding window. We generate 6000+ samples for each trajectory dataset using a 2 s spatial gap between sliding windows. We proceed similarly to generate more data with different sets of random input samples for each of the three traffic scenarios using a $5\%$ sampling rate. The final augmented dataset has $64000+$ input-output sample pairs for training the Deep CNN model. Note that the samples extracted from a specific trajectory record form a sequence, which violates the i.i.d assumption (independent and identically distributed) for the neural network training. However, this is rectified during the optimization stage, where only a random subset of the samples is used in each iteration of the CNN training (this is a common trick employed while training reinforcement learning models, for instance, the use of ``replay memory'' in \cite{mnih2015humanlevel}). We use the following additional parameters for training data generation: $|x|=10$ m, $|t|=1$ s, $c_w=18$ kmph, $c_v^{\mathrm{max}}=100$ kmph, $c_v^{\mathrm{min}}=60$ kmph, $V_\mathrm{max}=95$ kmph, $l_\mathrm{up}=80$ m and $l_\mathrm{dn}=40$ m.

We train five instances of the anisotropic and isotropic CNN models, and report the average of their performance results. We use the \textit{TensorFlow} framework \cite{tensorflow} to train all the models. The two major hyper-parameters, namely the CNN kernel width and depth in each layer, are independently optimized using the Hyperband algorithm \cite{Lisha2018hyperband}, which belongs to the class of bandit-based algorithms. Other hyper-parameter choices are: gradient descent batch size: 32 samples, total training epochs: 300, (fixed) learning rate: $1e-3$, and optimizer: Adam \cite{kingma2014adam}. We use a GPU cluster with NVIDIA Tesla V100 32GB for training the models. The run time for a single training experiment is between $120$ and $150$ min. Note that the training can be viewed as an offline procedure.

\subsection{Testing data}

We test our model using three datasets: (i) a hold-out set from the simulated data that is not used for training (from a different lane of the freeway section), (ii) the Next Generation Simulation Program (NGSIM) dataset \cite{ngsim}, and (iii) the German Highway Drone (HighD) dataset \cite{krajewski2018highd}. We choose the US-101 highway trajectory data from NGSIM, which contains the locations and speeds of all vehicles crossing the observed area during a $45$ min time period with a $0.1$ s resolution. The HighD data consists of trajectory data from several German highways, each consisting of a frame-wise recording of all vehicles passing a $400$ m section during a $20$ min duration, with a resolution of $25$ frames/second. The input-output test samples are generated similarly to the training datasets with the respective space-time discretization parameters. 

\emph{We emphasize that we do not use the NGSIM or HighD datasets for training the model.} In other words, the model is trained with data from a simulation using a freeway in the United Arab Emirates; and then tested with additional data from that same simulation, as well as real data from a freeway in the United States and several freeways in Germany. This allows us to evaluate the model's transferability to diverse traffic scenarios and dynamics not seen in the training set, and the viability of using simulated data instead of real data for training.

\section{Results and Discussion}
\label{sec4}

In this section, we present the anisotropic reconstruction results and compare the isotropic and anisotropic models. We also discuss the transferability of the trained models to real-world traffic conditions and extend the results to handle varying PV penetration rates.

The architecture of the CNN model obtained from the hyper-parameter optimization is shown in Table \ref{tab:model_arch}. We use the same optimized architecture for both the anisotropic and isotropic CNN models.

\begin{table}[!hbt]
	\centering
	\caption{Model architecture as obtained from the hyper-parameter optimization}
	\label{tab:model_arch}
	\begin{tabular}{@{}ccc@{}}
		\toprule
		Layer name & \multicolumn{1}{l}{Kernel widths} & \multicolumn{1}{l}{Kernel depths} \\ \midrule
		Conv-1    & ($5 \times 5$)                  & $40$                            \\
		Conv-2    & ($7 \times 7$)                  & $48$                            \\
		Conv-3    & ($7 \times 7$)                  & $32$                            \\
		Conv-4    & ($5 \times 5$)                  & $48$                            \\
		Conv-5    & ($5 \times 5$)                  & $40$                            \\
		Conv-6    & ($9 \times 9$)                  & $56$                            \\
		Output    & ($7 \times 7$)                  & $1$                             \\ \bottomrule
	\end{tabular}
\end{table}

\subsection{Anisotropic CNN model reconstruction}

\begin{figure*}[!hbt]
	\centering
	\resizebox{0.5\textwidth}{!}{%
		\includegraphics{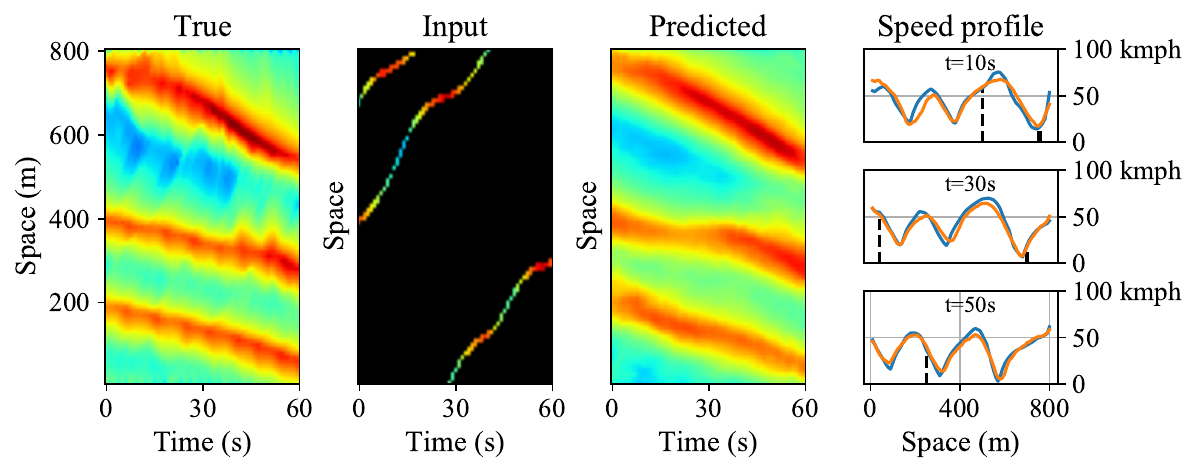}}
	
	{\small (a) Congested traffic, RMSE = $5.34$ kmph}
% 	\vspace{0.1in}
	
	\resizebox{0.5\textwidth}{!}{%
		\includegraphics{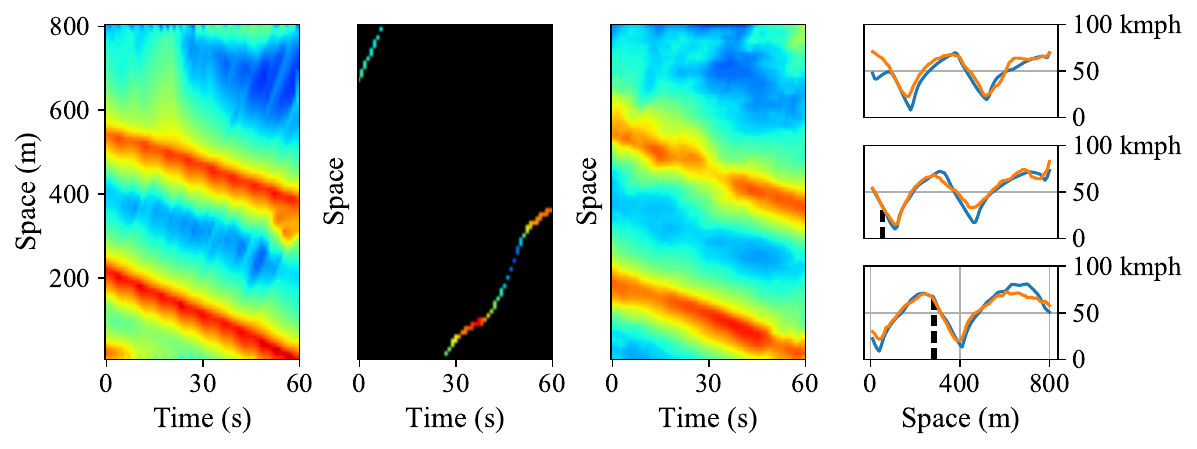}}
	
	{\small (b) Congested traffic, RMSE = $7.84$ kmph}
% 	\vspace{0.1in}
	
	\resizebox{0.5\textwidth}{!}{%
		\includegraphics{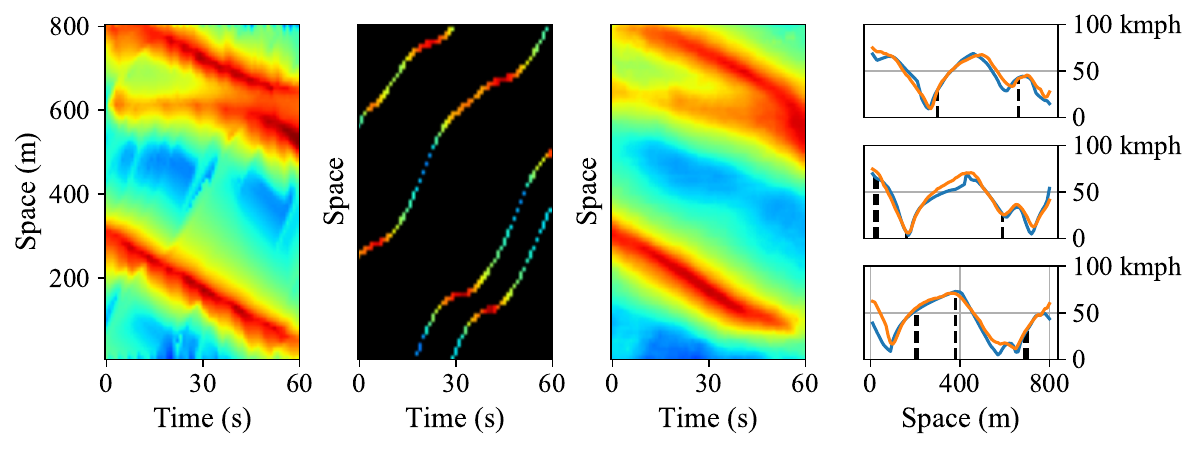}}
	
	{\small (c) Congested traffic, RMSE = $6.36$ kmph}
% 	\vspace{0.1in}
	
	\resizebox{0.5\textwidth}{!}{%
		\includegraphics{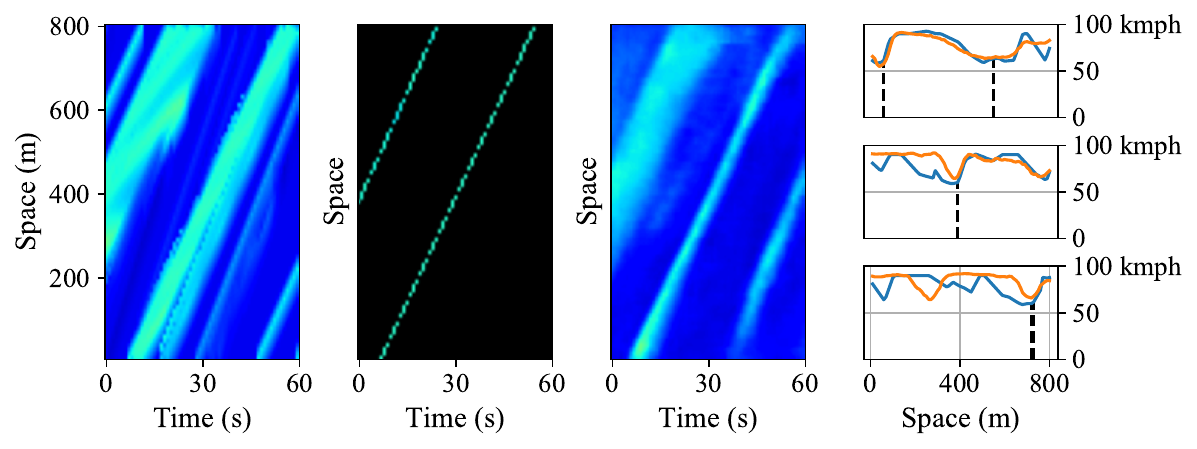}}
	
	{\small (d) Slow-moving traffic, RMSE = $10.78$ kmph}
% 	\vspace{0.1in}
	
	\resizebox{0.5\textwidth}{!}{%
		\includegraphics{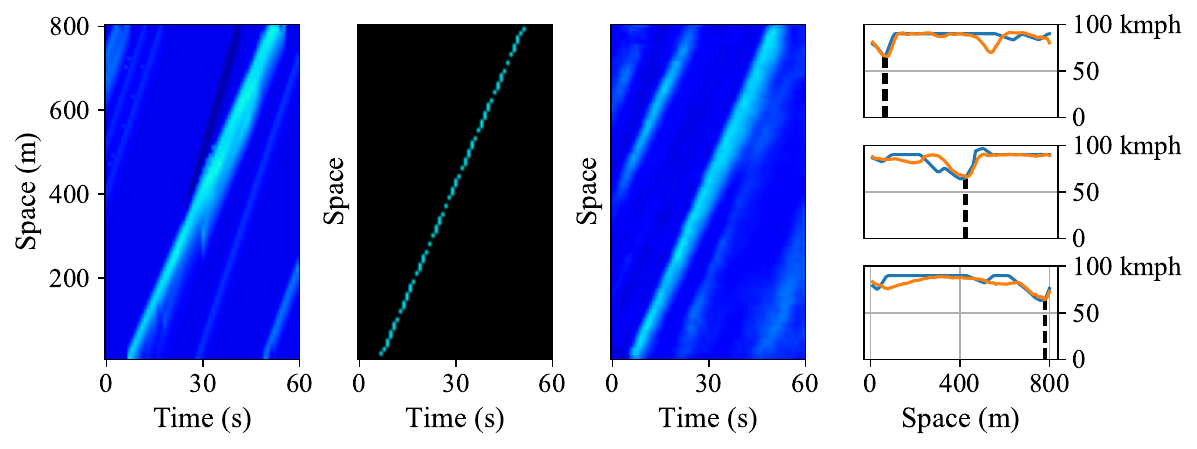}}
	
	{\small (e) Free-flowing traffic, RMSE = $5.54$ kmph}
	
	\caption{Estimated speed field for selected samples in the simulated test data using the anisotropic CNN model. The fourth column shows the speed profile across the road section at $t = 10, ~30$, and $50$ secs (\textit{blue - true speeds, orange - estimated speeds, black vertical dashed line - input PV speed}).}
	\label{fig:recon1}
\end{figure*}

Fig.~\ref{fig:recon1} shows five sample estimated speed fields from the hold-out simulated test dataset using the anisotropic model. The reconstruction window is $800$ m $\times$ $60$ s with a $10$ m $\times$ $1$~s resolution. The true speed field, PV trajectories, and speed profiles at three time instants ($t = 10, ~30, $ and $50$ s) are also shown for each sample. Three of the samples correspond to congested traffic conditions, one corresponds to slow-moving traffic conditions, and one corresponds to free-flowing traffic conditions.

There are several points of interest to note about the reconstruction: All the estimated speed fields are feasible in terms of traffic physics and capture the different traffic states well. The model reproduces the existence of free-flow, congested and transition traffic dynamics correctly despite having very limited input information from the PV trajectories. One can observe the accurate prediction of shockwave dynamics in the congested traffic samples (a)-(c). This is also evident from the speed profile comparison. The true speed profile is often noisy, and the reconstruction has a smoothing effect due to the local convolutional operations in the CNN layers.

We have observed that the estimated speeds in slow-moving traffic have a higher root mean squared error (RMSE) than those in congested and free-flowing traffic; see Fig.~\ref{fig:recon1} (d)-(e) and Table~\ref{tab:model_comp}. In slow-moving traffic, heterogeneity (caused by different vehicle characteristics, driving behaviors, etc.) is predominant, and one can see different forward wave propagation velocities in the speed field; see the example in Fig.~\ref{fig:recon1} (d). Therefore, estimation is inherently a challenging problem unless we observe the actual travel speed. This is not the case for congested traffic, where the collective dynamics can be inferred from the trajectory of a single vehicle, or for free-flowing traffic, where the traffic heterogeneity is limited. In short, traffic speed fields with varied forward propagation wave velocities are still difficult to infer. 
% \textcolor{blue}{Similar reasoning also applies in the case of traffic in the transition phase, which consists of shockwaves waves of varying speeds. Anisotropic kernels with restricted backward wave speeds could be challenging to capture these minor waves; however, these waves only persist for a short period and can be safely ignored in the estimation.} 
Interestingly, in all the scenarios, the anisotropic model predicts the average desired vehicle speed in areas where there are no PV trajectories, which is a reasonable conclusion when no vehicles are observed.

We interpret the Deep CNN model as an interpolation function that locally propagates traffic characteristics (forward and backward waves) using the sparse information from input vehicle trajectories. The model ensures sound propagation of traffic information in space and time, resulting in speed field estimates with different traffic states - free-flowing, slow-moving, congested, and their transition states. Introducing anisotropic kernels further limits the propagation speeds of traffic information, in accordance with the Kinematic Wave Theory of traffic flow. This results in speed field estimates with a gradual and physically reasonable transition between the different traffic states.  In contrast, traditional Kalman Filter based assimilation techniques only exploit state information from one (or a few) time step(s) when estimating the traffic speeds. This is inefficient in terms of data usage and fails to accurately reconstruct the dynamics. 

In addition, we have tuned the Deep CNN model architecture to learn different traffic wave dynamics. Whether to produce a backward or forward wave depends on the traffic regime, which the model infers from the input trajectories. This is confirmed from the latent space projection of the data (i.e., the output from the encoder model), where three distinct clusters were generated, corresponding to free-flowing, slow-moving and congested traffic, respectively. Another way to put this is that a neural network model can solve an under-determined system - a major upside compared to other machine learning models. This is in contrast with traditional estimation methods which require additional information on initial/boundary conditions or traffic demands.

\subsection{Comparison of anisotropic and isotropic models}

We next compare the performance and computational requirements of the anisotropic and isotropic models in Table~\ref{tab:model_comp}. The RMSE calculation shown in the table is the sample average for $4000+$ simulated test samples. Overall, the anisotropic and isotropic models have similar performance in terms of accuracy, but the anisotropic model leads to more physically plausible shockwave dynamics (this is discussed below). In particular, the anisotropic model performs slightly better in estimating the congested and free-flowing traffic in comparison to slow-moving traffic. This is because the slow-moving data samples comprise heterogeneous traffic in the free-flow regime, which might be better observed by an isotropic kernel than a restricted anisotropic kernel. Depending on the desired speed distribution, one can increase the extent of the anisotropic kernel (i.e, values of $c_v^{\rm min}$ and $c_w^{\rm max}$) and rectify this.

\begin{table}[!htb]
\centering
	\caption{Comparison of anisotropic and isotropic models. Percent change is with respect to isotropic model.}
	\label{tab:model_comp}
    \resizebox{\columnwidth}{!}{\begin{tabular}{@{}clccc@{}}
		\toprule
		\multicolumn{2}{c}{Metric} &
		\begin{tabular}[c]{@{}c@{}}Isotropic\\ model\end{tabular} &
		\begin{tabular}[c]{@{}c@{}}Anisotropic\\ model\end{tabular} &
		\begin{tabular}[c]{@{}c@{}}Percent\\ change\end{tabular} \\ \midrule
		\multirow{4}{*}{\begin{tabular}[c]{@{}c@{}}Root mean\\ squared error\\ (\textit{kmph})\end{tabular}} &
		Congested &
		\begin{tabular}[c]{@{}c@{}}$8.60$\\ $(\pm 3.16)$\end{tabular} &
		\begin{tabular}[c]{@{}c@{}}$8.50$\\ $(\pm 3.15)$\end{tabular} &
		$-1.2\% $ \\ \cmidrule(l){2-5} 
		&
		Slow-moving &
		\begin{tabular}[c]{@{}c@{}}$10.37$\\ $(\pm1.60)$\end{tabular} &
		\begin{tabular}[c]{@{}c@{}}$10.53$\\ $(\pm1.70)$\end{tabular} &
		$+1.5\% $ \\ \cmidrule(l){2-5} 
		&
		Free-flowing &
		\begin{tabular}[c]{@{}c@{}}$7.42$\\ $(\pm 2.23)$\end{tabular} &
		\begin{tabular}[c]{@{}c@{}}$7.40$\\ $(\pm2.22)$\end{tabular} &
		$-0.3\% $ \\ \cmidrule(l){2-5} 
		&
		Total &
		\begin{tabular}[c]{@{}c@{}}$8.71$\\ $(\pm 2.76)$\end{tabular} &
		\begin{tabular}[c]{@{}c@{}}$8.76$\\ $(\pm 2.71)$\end{tabular} &
		$+0.5\% $ \\ \midrule
		\multicolumn{2}{c}{Number of parameters} &
		$443193$ &
		$215625$ &
		$-51.4\% $ \\ \bottomrule
	\end{tabular}}
\end{table}

Table~\ref{tab:model_comp} also shows that the anisotropic model requires only half as many parameters as the isotropic model, which is a significant improvement in model complexity given that the performance of the two models is very similar. From a computational perspective, this is a substantial advantage, leading to faster model convergence (RMSE reduction per training epoch) and a potential reduction in the number of training samples required. This confirms that exploiting domain knowledge results in simpler and more interpretable learning models.

\begin{figure}[!hbt]
	\centering
	\resizebox{0.4\textwidth}{!}{%
		\includegraphics{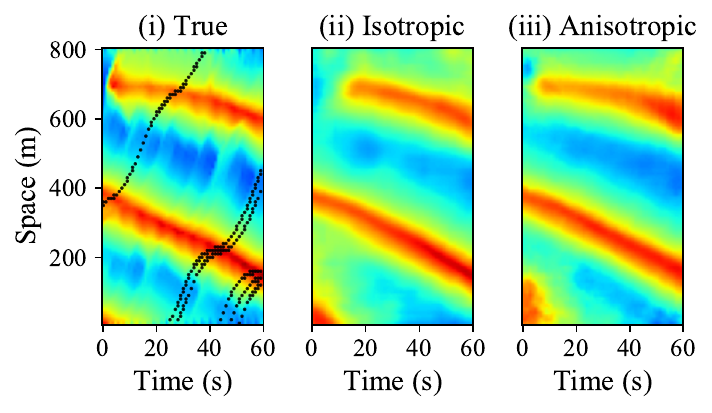}}
	
	{\small (a) Example 1 (Congested traffic, RMSE = $5.34$ kmph)}
	\vspace{0.1in}
	
	\resizebox{0.4\textwidth}{!}{%
		\includegraphics{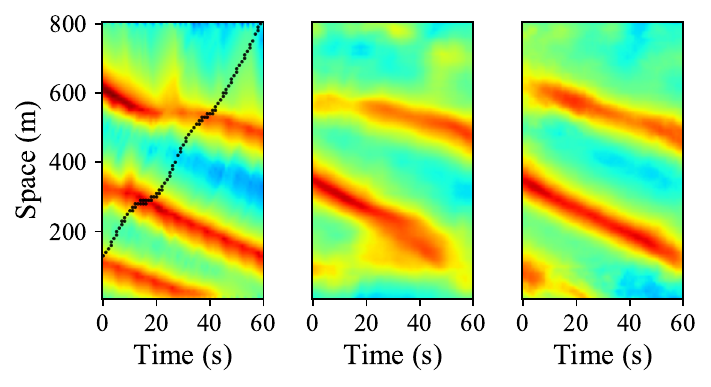}}
	
	{\small (b) Example 2 (Congested traffic, RMSE = $7.84$ kmph)}
	\vspace{0.1in}
	
	\resizebox{0.4\textwidth}{!}{%
		\includegraphics{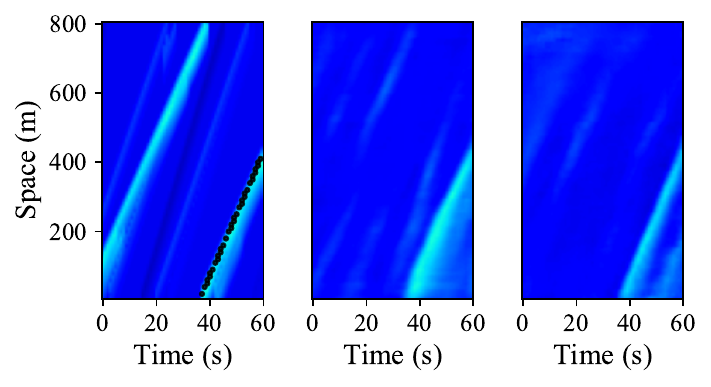}}
	
	{\small (c) Example 3 (Free-flowing traffic, RMSE = $6.36$ kmph)}
	\vspace{0.1in}
	
	\caption{Estimated speed fields for some selected samples in the simulated test data using the anisotropic and the isotropic CNN models. Black lines show the probe vehicle trajectories used for the reconstruction.}
	\label{fig:recon-comp}
\end{figure}

Although the isotropic and anisotropic models perform comparably in terms of the average error in estimating the speed, there are some examples where they differ in terms of the structure (speed and extent) of the shockwaves they produce. This is illustrated in Fig.~\ref{fig:recon-comp}, which shows certain examples where the anisotropic model clearly reconstructs more physically plausible shockwave dynamics, as mentioned below.

In the example in Fig. \ref{fig:recon-comp}(a), the isotropic CNN underestimates the length of the shockwave at the top, whereas the anisotropic CNN correctly predicts that it existed some time prior to the two PV trajectories crossing it. This is because the anisotropic kernel gets more activation along the direction of the shockwave and hence reconstructs the stop-and-go region correctly, whereas the isotropic kernel considers all directions, which possibly results in averaging out all the nearby activations. Similar patterns have been observed in other test instances. The estimation in Fig. \ref{fig:recon-comp}(b) is obtained using a single input trajectory. The anisotropic model gives a plausible reconstruction of the shockwave whereas the isotropic reconstruction shows large dispersion, which is also physically inconsistent with the input data. The design of anisotropic kernels can rule out such inconsistencies arising in the estimation. Fig. \ref{fig:recon-comp}(c) shows a free-flowing traffic estimation. Again the forward wave produced by the isotropic kernel has more dispersion. In summary, one can see that the anisotropic model produces more accurate wave propagation dynamics consistent with traffic physics, even though the RMSEs of the models are similar.

\begin{figure}[hbt!]
    \begin{center}
        \includegraphics[width=0.23\textwidth]{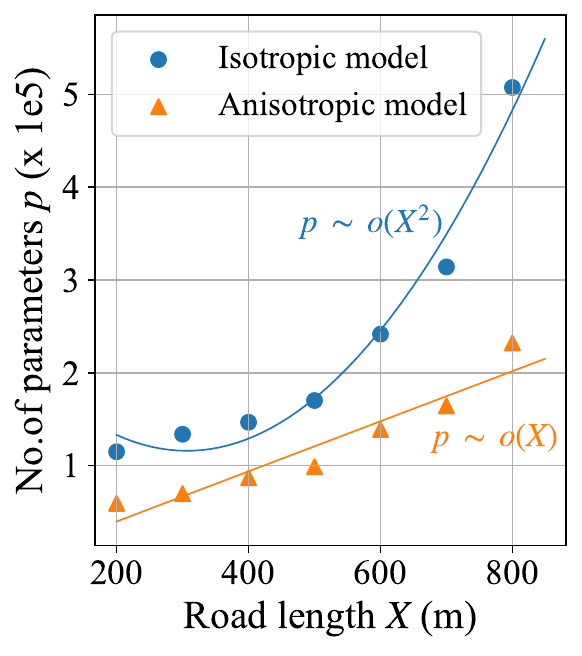}%
        \includegraphics[width=0.23\textwidth]{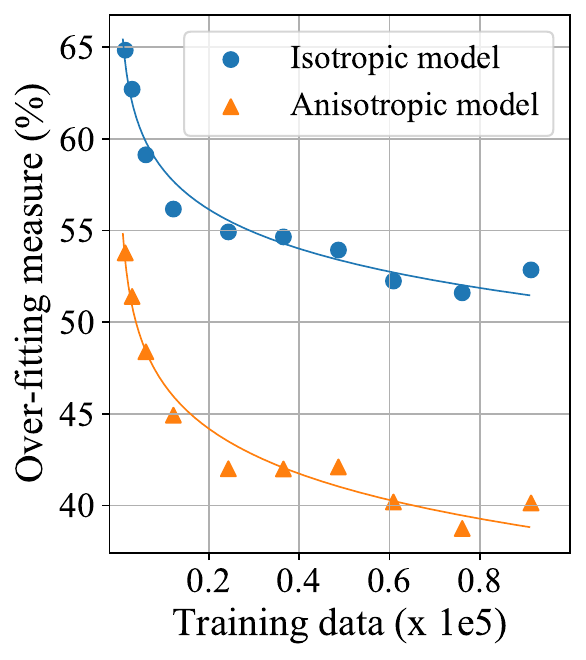}
        \newline {\small (a) Model complexity \qquad (b) Over-fitting measure}
    \end{center}
    \caption{Comparison of model-complexity and over-fitting measures for isotropic and anisotropic CNN models.}%
    \label{fig:modelcomplx}%
\end{figure}

We finally compare the anisotropic and isotropic models from the perspective of model complexity and over-fitting measures. In order to understand how the complexity of CNN models scales with the road network size, we optimize the anisotropic and isotropic CNN model architectures for different road lengths. The optimization is done using the Hyperband algorithm \cite{Lisha2018hyperband}. The results are shown in Fig.~\ref{fig:modelcomplx}(a), where the optimal number of model parameters required for different road lengths are compared (see the scatter plot). We see that the model complexity scales \emph{quadratically} for isotropic kernels, whereas for anisotropic models it scales \emph{linearly} (see the curve plot). Thus, as the problem size becomes large (for e.g., for long road sections, multiple lanes, or network level settings), the optimal CNN model architecture required for learning traffic dynamics becomes significantly large with isotropic kernels. The anisotropic CNN model, on the other hand, scales well to large problem sizes, results in simpler and more manageable models, and is beneficial for practical implementation. This observation is in-line with the scaling results obtained in Fig.~\ref{fig:kersize}.

In order to measure the over-fitting in CNN models, we define an over-fitting metric as,
\begin{equation}
O_{\rm fit} = \Bigg| \frac{{\rm RMSE}_{\rm~train} - {\rm RMSE}_{\rm~test}}{{\rm RMSE}_{\rm~test}} \Bigg| \times 100,
\end{equation}
where ${\rm RMSE}_{\rm~train}$ and ${\rm RMSE}_{\rm~test}$ are the root mean squared error metrics for training and testing data respectively. $O_{\rm fit}$ measures the difference in the model's performance on the training and testing data; higher values for $O_{\rm fit}$ imply more over-fitting. Over-fitting is an undesirable feature, and indicates poor generalization to unseen testing data. Fig.~\ref{fig:modelcomplx}(b) shows $O_{\rm fit}$ for the isotropic and anisotropic CNN models trained with different proportions of the total training data. The simulated training and simulated testing data are used to calculate $O_{\rm fit}$. Note that the hyper-parameters of the CNN models are optimized independently to ensure that $O_{\rm fit}$ is compared for the optimal isotropic and anisotropic models. The trend line in Fig.~\ref{fig:modelcomplx}(b) shows that the isotropic CNN model results in higher over-fitting. Since this observation is consistent at all data levels (and thus independent of model complexity), we conclude that the isotropic model has higher tendency to over-fit than the proposed anisotropic model. This is because the anisotropic CNN model reduces the number of parameters in a principled way, which lowers the model complexity without compromising test accuracy. In other words, the introduction of anisotropic kernels is a natural way to train CNN models that learn traffic speed dynamics with a lowered risk of over-fitting.

\subsection{Transferability to real-world traffic dynamics}

\begin{figure}[!hbt]
	\centering
	\includegraphics[width=0.45\textwidth]{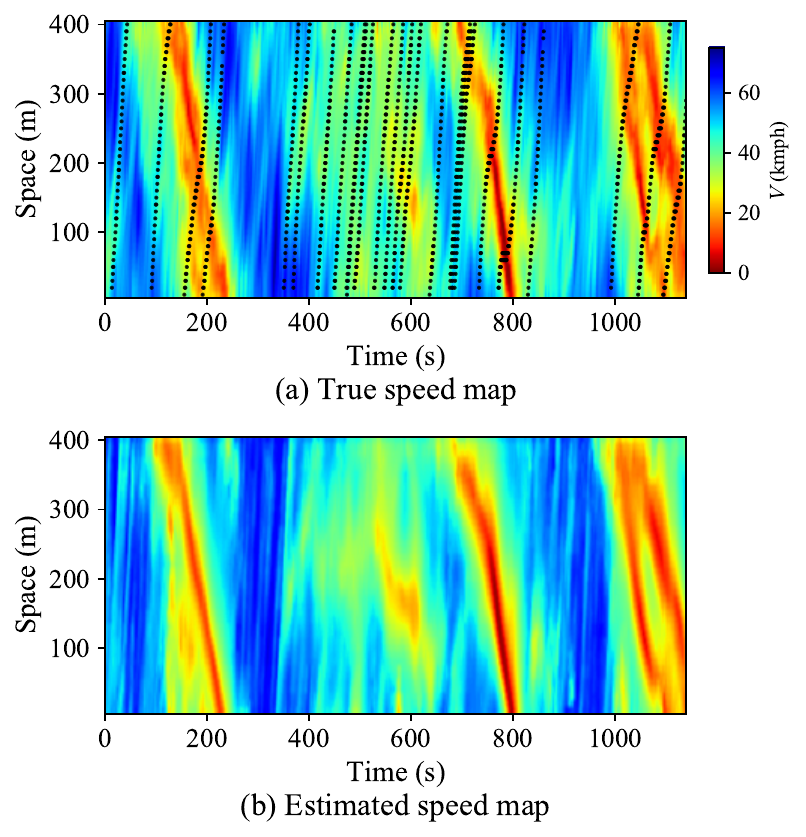}
	\caption{Estimated speed field of lane 4 of highway No. 25 in the HighD dataset using $5\%$ probe sampling rate. The road section is $X=400$ m long and the reconstruction period is $T=1140$ s. The RMSE is $6.80$ kmph.}
	\label{fig:highd-hw25}
\end{figure}

\begin{figure}[!hbt]
	\centering
	\includegraphics[width=0.45\textwidth]{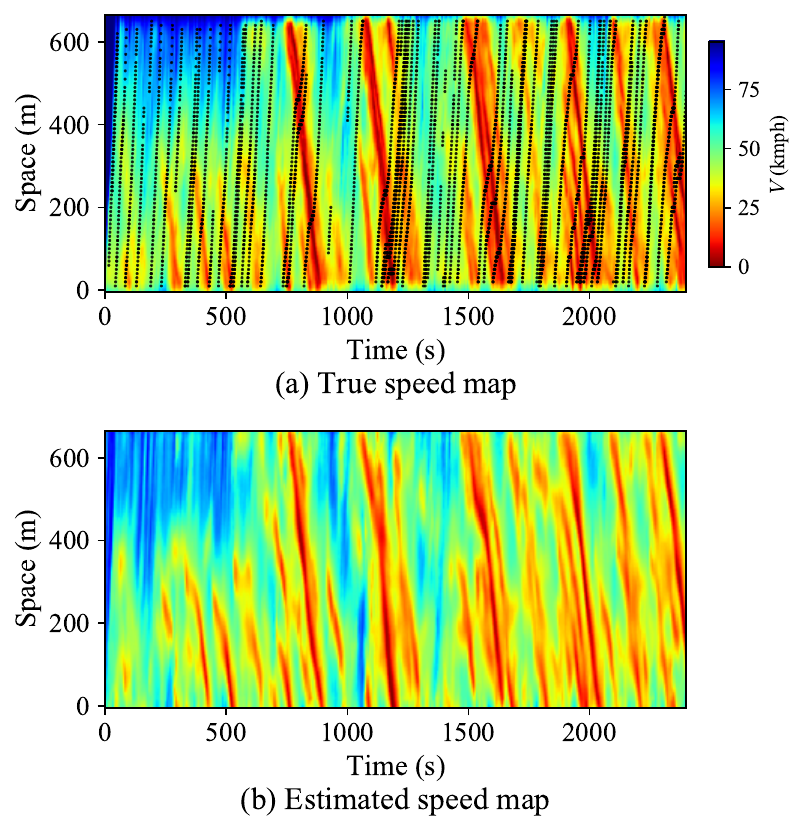}
	\caption{Estimated speed field of lane 2 of U.S. Highway 101 in the NGSIM dataset using $5\%$ probe sampling rate. The road section is $X=670$ m long and the reconstruction period is $T=2400$ s. The RMSE is $10.50$ kmph.}
	\label{fig:ngsim}
\end{figure}

\begin{figure}[!hbt]
	\centering
	\includegraphics[width=0.45\textwidth]{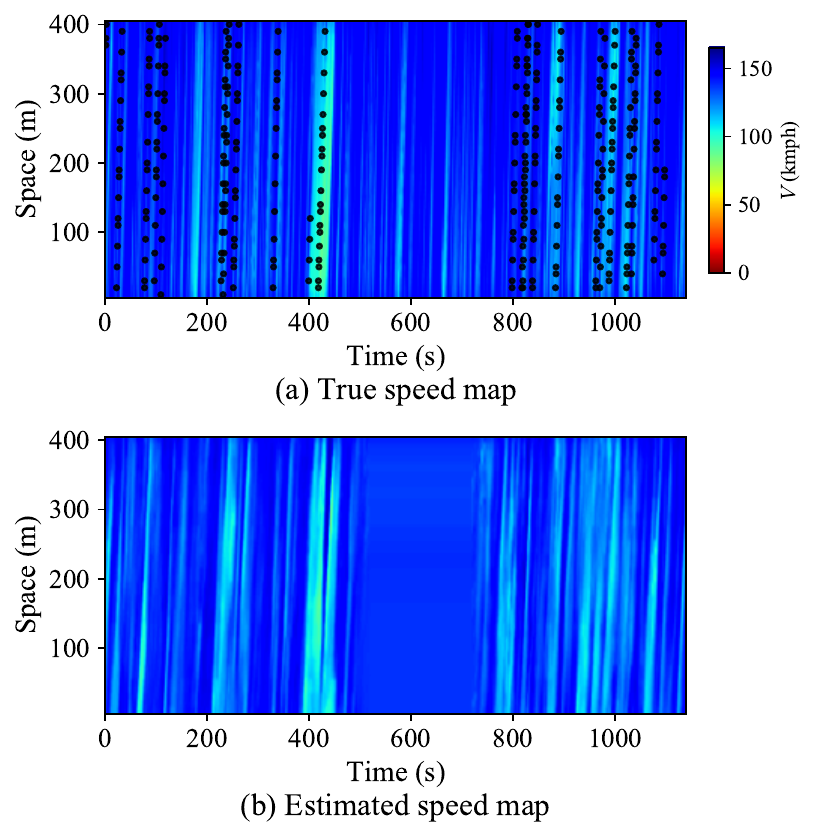}
	\caption{Estimated speed field of lane 6 of Highway No. 44 in the HighD dataset using $5\%$ probe sampling rate. The road section is $X=400$ m long and the reconstruction period is $T=1140$ s. The RMSE is $14.60$ kmph.}
	\label{fig:highd-hw44}
\end{figure}

To understand how well the anisotropic CNN model performs in scenarios with different traffic characteristics than those observed in the training dataset, we test it on various real-world freeway sections. Figs. \ref{fig:highd-hw25} to \ref{fig:highd-hw44} show the estimation results for three sample freeway sections from the HighD and NGSIM datasets using data with a PV sampling rate of $5\%$.

A quick observation shows that all three example reconstructions are plausible, despite having different space-time dimensions from those used in the training dataset. This is possible because of the parameter sharing property of CNNs, whereby the features learned during training (traffic characteristics in this case) are space-time invariant, and hence can be used with any spatio-temporal reconstruction window.

Closer observation reveals variation in the performance across the three examples. The estimated speed field in Fig.~\ref{fig:highd-hw25} has the lowest RMSE ($\approx 6.80$ kmph), and the speed, width and duration of the predicted backward propagating shockwaves are accurate. In the estimation in Fig.~\ref{fig:ngsim}, the shockwave reconstruction and speeds are reasonably correct, and the RMSE is moderate ($\approx 10.50$ kmph). The model correctly predicts the onset of shockwaves $400$ m upstream of the road section during the initial 600 seconds, though it underestimate the shockwave width. Thus, the model can accurately estimates speed fields for a road section that is simultaneously congested and free-flowing during the same period. The estimation result in Fig.~\ref{fig:highd-hw25} also supports this. The third freeway section, shown in Fig.~\ref{fig:highd-hw44}, comprises free-flowing traffic and has the highest RMSE ($\approx 14.60$ kmph). Apart from the inherent difficulties in the estimation of free-flowing traffic speeds, one can also observe that the speed of forward 
waves predicted by the model is slightly lower than that from the true waves; see the slow-moving band around $400$ secs as an example. This is due to the difference in the traffic characteristics in the training and testing data, which we elaborate on below.

\begin{figure*}[hbt!]
	\centering
	\includegraphics[width=0.24\textwidth]{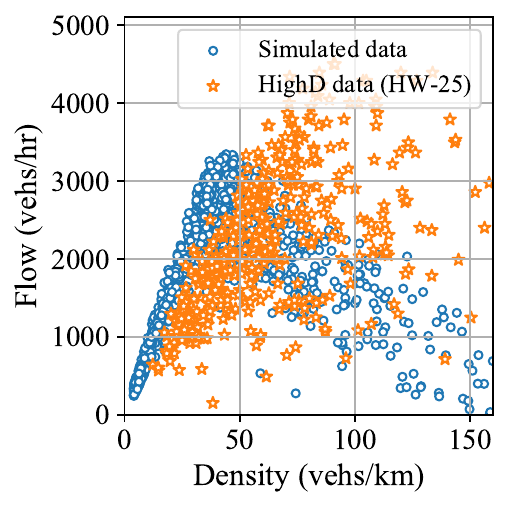} \hspace{0.2in}
	\includegraphics[width=0.24\textwidth]{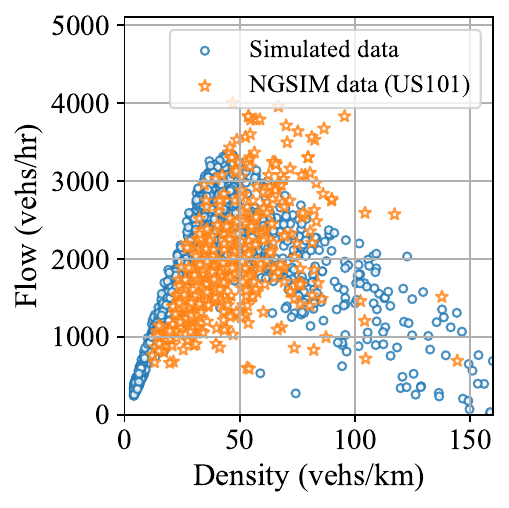} \hspace{0.2in}
	\includegraphics[width=0.24\textwidth]{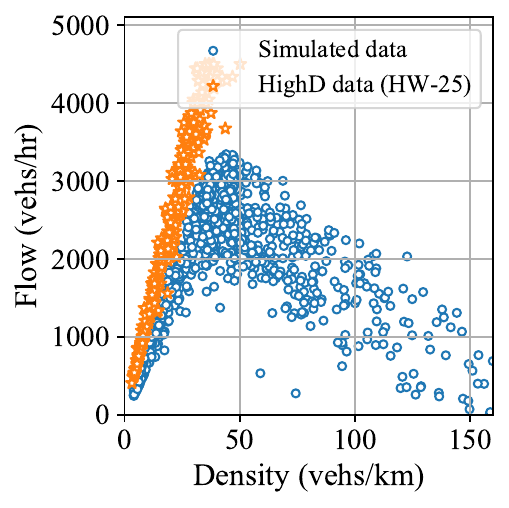}
	\newline
	{\small (a) HighD Highway 25 lane 4 \hspace{0.1in} (b) NGSIM US Highway 101 lane 2 \hspace{0.1in} (c) HighD Highway 44 lane 6}
	\caption{Flow-density scatter plot comparing the traffic characteristics of the real-world datasets with the simulated training data}
	\label{fig:data_comp}
\end{figure*}

Recall that the CNN model trained on the simulated data encompasses the knowledge of traffic dynamics of a specific freeway section. How well this model transfers to other test scenarios depends on the traffic characteristics of the test segment. One can explain the difference in the RMSE errors in Figs.~\ref{fig:highd-hw25} to \ref{fig:highd-hw44} by comparing the dynamics contained in the simulated data and the test data. One useful tool for this comparison is the flow-density scatter plot, which is shown in Fig.~\ref{fig:data_comp}. The reason for the low RMSE in the first two examples is that the freeways are operating in the congested regime and the shockwave speeds in the simulated and test data are similar. Likewise, the reason for slightly lower prediction of free-flowing speed in the third freeway section (which operates in free-flowing traffic) is evident from Fig.~\ref{fig:data_comp}(c). A similar reasoning is applicable when one discusses the transferability of the simulated section with its own real-world section, as the simulation doesn't capture the complete dynamics. Empirical FD comparisons can be further exploited to calibrate the trained deep learning model to match with the traffic dynamics of the test data. This is beyond the scope of the current work and represents a possible future extension.

In short, we observe that the Deep CNN models trained with simulated data from minimally calibrated traffic flow models transfer well to real-world traffic scenarios. Better results are to be expected with simulation models that are calibrated to the testing scenarios. Since the data required for calibration is lower than that needed for training deep neural networks, we can take advantage of well-developed microscopic traffic simulations to fit data-hungry models like CNNs.

We finally compare the anisotropic Deep CNN model performance with two other existing traffic speed estimation techniques in the literature: (a) General Adaptive Smoothing Method (GASM) from \cite{trieber2011filter}, and (b) Velocity-based LWR Ensemble Kalman Filtering technique (LWR-v EnKF) from \cite{daniel2008enkf}. While both techniques directly estimate the speed field over a given time-space plane, the former is a data assimilation technique using a macroscopic traffic flow model and the latter is an informed traffic interpolation procedure. The LWR-v EnKF method additionally requires initial and boundary conditions as inputs. The estimation results for the NGSIM US-101 highway lane 2 using the anisotropic CNN model, GASM and LWR-v EnKF are compared in Table~\ref{tab:est_comp}. The RMSE metric is evaluated for different input PV penetration rates to understand how well the techniques perform in sparse observation setting. Overall, we see that the anisotropic Deep CNN model results in the least estimation error for all PV penetration rates.

\begin{table}[hbt!]
\centering
\caption{Comparison results to existing estimation technqiues.}
\label{tab:est_comp}
\resizebox{\columnwidth}{!}{\begin{tabular}{@{}cccc@{}}
\toprule
\multirow{2}{*}{Estimation techniques} & \multicolumn{3}{c}{\begin{tabular}[c]{@{}c@{}}Root mean squared error (\textit{kmph}) \\ at different PV penetration rates\end{tabular}} \\ \cmidrule(l){2-4} 
 & $3\%$ & $5\%$ & $10\%$ \\ \midrule
\begin{tabular}[c]{@{}c@{}}Aniso CNN model\\ (this paper)\end{tabular} & \begin{tabular}[c]{@{}c@{}}$11.60$\\ $(\pm 1.46)$\end{tabular} & \begin{tabular}[c]{@{}c@{}}$10.70$\\ $(\pm 0.59)$\end{tabular} & \begin{tabular}[c]{@{}c@{}}$8.88$\\ $(\pm 0.24)$\end{tabular} \\[0.4cm]
\begin{tabular}[c]{@{}c@{}}GASM method\\ (Treiber et al. 2011 \cite{trieber2011filter})\end{tabular} & \begin{tabular}[c]{@{}c@{}}$13.49$\\ $(\pm 2.50)$\end{tabular} & \begin{tabular}[c]{@{}c@{}}$11.80$\\ $(\pm 1.90)$\end{tabular} & \begin{tabular}[c]{@{}c@{}}$9.50$\\ $(\pm 1.06)$\end{tabular} \\[0.4cm]
\begin{tabular}[c]{@{}c@{}}LWR-v EnKF method\\ (Daniel et al. 2008 \cite{daniel2008enkf})\end{tabular} & \begin{tabular}[c]{@{}c@{}}$14.93$\\ $(\pm 0.05)$\end{tabular} & \begin{tabular}[c]{@{}c@{}}$14.64$\\ $(\pm 0.08)$\end{tabular} & \begin{tabular}[c]{@{}c@{}}$13.63$\\ $(\pm 0.20)$\end{tabular} \\ \bottomrule
\end{tabular}}
\end{table}

We found that the GASM method provides reasonable estimates at higher PV penetration rates (for e.g., $\geq 10\%$). However, at lower PV penetration rates (i.e., when the input only consist of one or two PV trajectories), the GASM method fails to reproduce correct traffic speed waves and results in higher estimation error. We have also noticed that the GASM method produces large dispersion in their estimates, which implies it poorly captures short-term traffic variations and is not suitable for high-resolution estimation. This is because the GASM method only uses two (pre-defined) kernels for interpolation, while our anisotropic CNN model uses an ensemble of (learned) kernels and hence interpolates low-level traffic features well.

Traffic speed estimated using the LWR-v EnKF method results in highest RMSE as shown in Table~\ref{tab:est_comp}. This poor performance could be due to its myopic character $-$ only uses traffic speed inputs at the current time steps whereas the anisotropic CNN model considers inputs from multiple time steps. We also notice that the performance gets worse for longer estimation intervals, since the LWR model predictions deviate significantly from the actual data. Similar to GASM, the LWR-v EnKF also captures macroscopic traffic features and gets better at higher PV penetration rates.

In short, the anisotropic CNN model outperforms the existing traffic speed estimation techniques, especially at lower PV penetration rates.

\subsection{Variable probe vehicle penetration rates}

\begin{figure}[hbt!]
    \begin{center}
        \includegraphics[width=0.23\textwidth]{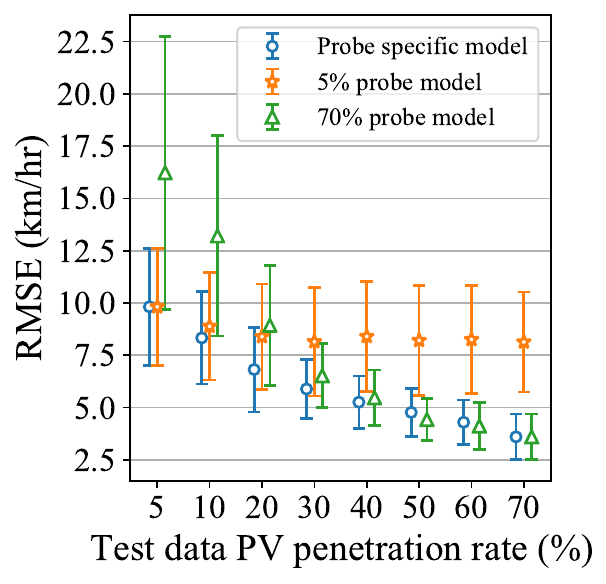}%
        \includegraphics[width=0.23\textwidth]{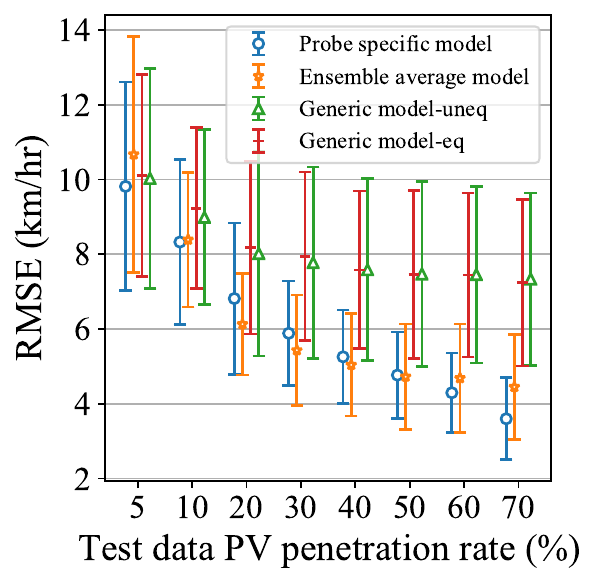}
        \newline {\small (a) Probe specific models \qquad (b) Generic models}
    \end{center}
	\caption{Probe vehicle (PV) penetration rate analysis}
	\label{fig:probe_res}
\end{figure}

To conclude our evaluation of the CNN model's performance, we investigate the effect of changing the PV penetration rate. We train six separate models using data consisting of specific PV penetration rates $10\%, ~20\%, ~\dots, ~70\%$ respectively (in addition to the \textit{5\% probe model} discussed so far in this paper). The input-output pairs for training are generated in the same way as explained in  Sec.~\ref{sec3} B. Each of these \textit{probe specific models} is evaluated using testing data which has a corresponding PV penetration rate to the respective model. The average test RMSE results are shown in Fig.~\ref{fig:probe_res}(a) (labeled "probe specific model"). As expected, the RMSE decreases with higher PV penetration rates.

However, we find that these probe specific models are not trivially generalizable to handle penetration rates other than what they were trained on, i.e., the models are penetration rate dependent. We demonstrate this by evaluating the performance of the two extreme probe specific models (i.e. the $5\%$ probe model and the $70\%$ probe model) on testing data across the whole range of PV penetration rates. It is clear that these probe specific models perform well only in/near their training domain. This could be due to the unconstrained latent space representation while training the CNN models, and is inevitable in any data driven models unless physical constraints are imposed.

% \begin{figure}[!h]
% 	\centering
% 	\includegraphics[width=0.35\textwidth]{Figures/Probe_analysis1.pdf}
	
% 	{\small (a) Probe specific models}
% 	\vspace{0.2in}
	
% 	\includegraphics[width=0.35\textwidth]{Figures/Probe_analysis.pdf}
	
% 	{\small (b) Generic models}
	
% 	\caption{Probe vehicle penetration analysis}
% 	\label{fig:probe_res}
% \end{figure}

The actual PV penetration rate depends on the prevailing traffic demand on the freeway, which is hard to measure in practice. We aim for an estimation model that performs well irrespective of the PV penetration rate. In other words, we want an estimation model that doesn't require prior knowledge of the PV penetration rate. Therefore, we test three methods to handle varying PV penetration rates. The first two methods are brute force approaches, which consist of training the CNN model on a dataset containing the whole range of PV penetration rates $5\%, ~10\%, ~20\%, ~\dots, ~70\%$. The third method is to use an ensemble CNN model. The RMSE results from these models are compared in Fig.~\ref{fig:probe_res}(b).

The first model (labeled \textit{generic model-eq}) is trained on a dataset consisting of all PV penetration rates sampled in equal proportion. The second model (labeled \textit{generic model-uneq}) is similar except that we give more importance to lower PV penetration rates which are more difficult to learn. This is achieved by including more data samples for lower PV penetration rates, i.e., training data $\propto$ 1/(PV penetration rate). Both these models, however, perform sub-optimally compared to the probe specific models. The best method is the third model (labeled \textit{ensemble average model}), which takes the average of the predictions of all the probe specific models. This is referred to as “ensemble bagging” in the machine learning literature, and performs better than a single model trained on a wide range of penetration rates.  As seen in Fig.~\ref{fig:probe_res}(b), the ensemble CNN performs consistently well across all the PV penetration rates, even outperforming the respective probe specific models in certain cases. In addition to the performance, the individual models in the ensemble CNN (also called weak learners) can be trained in parallel, resulting in significantly lower training time than the other two generic models.

\section{Conclusion}
\label{sec5}

Deep learning models have shown success in solving several inverse problems in traffic flow, but they are limited by their lack of robustness and poor model interpretability. In this paper, we overcome these limitations by proposing an anisotropic Deep Convolutional Neural Network (CNN) model for estimating high-resolution traffic speed field using measurements from probe vehicles. The model employs anisotropic traffic kernels which are designed to explicitly capture a broad range of forward and backward propagation speeds in macroscopic traffic. Additionally, the Deep CNN model is trained using simulated traffic data. Since the generalization of Deep CNN performance depends on the distribution of training data, we note that using a targeted simulated data is an alternate method of imposing desirable traffic physics on the estimation model. For instance, we generate data corresponding to different traffic conditions (congested, slow-moving, free-flowing, etc.) so that the Deep CNN can learn traffic wave propagation speeds originating in heterogeneous traffic.

We present estimation results with input PV penetration rates as low as $5\%$ and output resolution as high as $10$m $\times$ $1$s. In the experiments, we primarily focused on the benefits of using anisotropic kernels in the Deep CNN model over the na\'ive isotropic kernels. We found that anisotropic kernels result in parsimonious model complexity and are less prone to model over-fitting, although the estimation error is similar to their isotropic counterparts. The model complexity grows linearly with problem size for anisotropic kernels whereas it grows as quadratic for isotropic kernels. Specific examples are provided to demonstrate that the anisotropic kernels better produce physically correct traffic shockwaves. We further evaluated the anisotropic Deep CNN on real-world traffic datasets and found acceptable transferability performance. This suggests that simulated data is a viable surrogate to real-world data for training Deep CNNs. We also found that the Deep CNN model performance is PV penetration rate dependent and proposed an ensemble model to handle unknown PV penetration rates.

We believe that the optimal way to apply learning techniques to a specific domain such as traffic state estimation is to integrate the fundamental principles of the domain into the framework of the learning model. This paper represents only one possible example of this general approach. In future work, we aim to explore other methods to incorporate traffic flow theory into learning models such as CNNs.

\section*{Acknowledgment}
This work was supported by the NYUAD Center for Interacting Urban Networks (CITIES), funded by Tamkeen under the NYUAD Research Institute Award CG001. The views expressed in this article are those of the authors and do not reflect the opinions of CITIES or its funding agencies.

\appendix

\bibliographystyle{plainnat}
\bibliography{references}
	
	
	
	
	
	
	% \end{thebibliography}

\end{document}